\def\year{2022}\relax
\documentclass[letterpaper]{article} 
\usepackage{aaai22}  
\usepackage{times}  
\usepackage{helvet}  
\usepackage{courier}  
\usepackage[hyphens]{url}  
\usepackage{graphicx} 
\usepackage{natbib}  
\usepackage{caption} 
\usepackage{amsmath}
\usepackage{amssymb}
\usepackage{xspace}
\usepackage{makecell}
\usepackage{multirow}
\usepackage{subfigure}

\usepackage[ruled,vlined]{algorithm2e}

\usepackage{xcolor}
\definecolor{bb}{rgb}{0.0, 0.0, 0.5}
\definecolor{deeppink}{rgb}{1.0, 0.08, 0.58}
\usepackage[colorlinks=true,linkcolor=red,citecolor=bb,urlcolor=deeppink]{hyperref}

\makeatletter
\DeclareRobustCommand\onedot{\futurelet\@let@token\@onedot}
\def\@onedot{\ifx\@let@token.\else.\null\fi\xspace}

\def\eg{\emph{e.g}\onedot} 
\def\ie{\emph{i.e}\onedot} 
 
\def\etc{\emph{etc}\onedot}

\makeatother

\DeclareMathOperator*{\argmax}{arg\,max}

\DeclareCaptionStyle{ruled}{labelfont=normalfont,labelsep=colon,strut=off} 
\usepackage{newfloat}
\usepackage{listings}
\usepackage{stfloats}
\title{BM-NAS: Bilevel Multimodal Neural Architecture Search}
\author{
    Yihang Yin\textsuperscript{\rm 1}, 
    Siyu Huang\textsuperscript{\rm 2}, 
    Xiang Zhang\textsuperscript{\rm 3}
}
\affiliations{
    \textsuperscript{\rm 1}Nanyang Technological University, \textsuperscript{\rm 2}Harvard University, \textsuperscript{\rm 3}The Pennsylvania State University\\
    yyin009@e.ntu.edu.sg, huang@seas.harvard.edu, xzz89@psu.edu
}

\usepackage{bibentry}

\begin{document}

\maketitle

\begin{abstract}
Deep neural networks (DNNs) have shown superior performances on various multimodal learning problems. However, it often requires huge efforts to adapt DNNs to individual multimodal tasks by manually engineering unimodal features and designing multimodal feature fusion strategies. This paper proposes Bilevel Multimodal Neural Architecture Search (BM-NAS) framework, which makes the architecture of multimodal fusion models fully searchable via a bilevel searching scheme. At the upper level, BM-NAS selects the inter/intra-modal feature pairs from the pretrained unimodal backbones. At the lower level, BM-NAS learns the fusion strategy for each feature pair, which is a combination of predefined primitive operations. The primitive operations are elaborately designed and they can be flexibly combined to accommodate various effective feature fusion modules such as multi-head attention (Transformer) and Attention on Attention (AoA). Experimental results on three multimodal tasks demonstrate the effectiveness and efficiency of the proposed BM-NAS framework. BM-NAS achieves competitive performances with much less search time and fewer model parameters in comparison with the existing generalized multimodal NAS methods. Our code is available at \url{https://github.com/Somedaywilldo/BM-NAS}.
\end{abstract}

\section{Introduction}

Deep neural networks (DNNs) have achieved a great success on various unimodal tasks (\eg, image categorization \cite{alexnet,he2016resnet}, language modeling \cite{vaswani2017attention,devlin2018bert}, and speech recognition \cite{amodei2016deepspeech2}) as well as the multimodal tasks (\eg, action recognition \cite{simonyan2014two,vielzeuf2018centralnet}, image/video captioning \cite{you2016captioning,jin2019captioning,jin2020captioning}, visual question answering \cite{lu2016vqa,anderson2018vqa}, and cross-modal generation \cite{reed2016generative, zhou2019generation}). Despite the superior performances achieved by DNNs on these tasks, it usually requires huge efforts to adapt DNNs to the specific tasks. Especially with the increase of modalities, it is exhausting to manually design the backbone architectures and the feature fusion strategies. It raises urgent concerns about the automatic design of multimodal DNNs with minimal human interventions.  

Neural architecture search (NAS) \cite{zoph2017nas,liu2018nas} is a promising data-driven solution to this concern by searching for the optimal neural network architecture from a predefined space. By applying NAS to multimodal learning, MMnas \cite{yu2020mmnas} searches the architecture of Transformer model for visual-text alignment and MMIF \cite{peng2020mmif} searches the optimal CNNs structure to extract multi-modality image features for tomography. These methods lack generalization ability since they are designed for models on specific modalities. MFAS \cite{perez2019mfas} is a more generalized framework which searches the feature fusion strategy based on the unimodal features. However, MFAS only allows fusion of inter-modal features, and the fusion operations are not searchable. It results in a limited space of feature fusion strategies when dealing with various modalities in different multimodal tasks.

\begin{figure}[t]
\begin{center}
\includegraphics[width=1\linewidth]{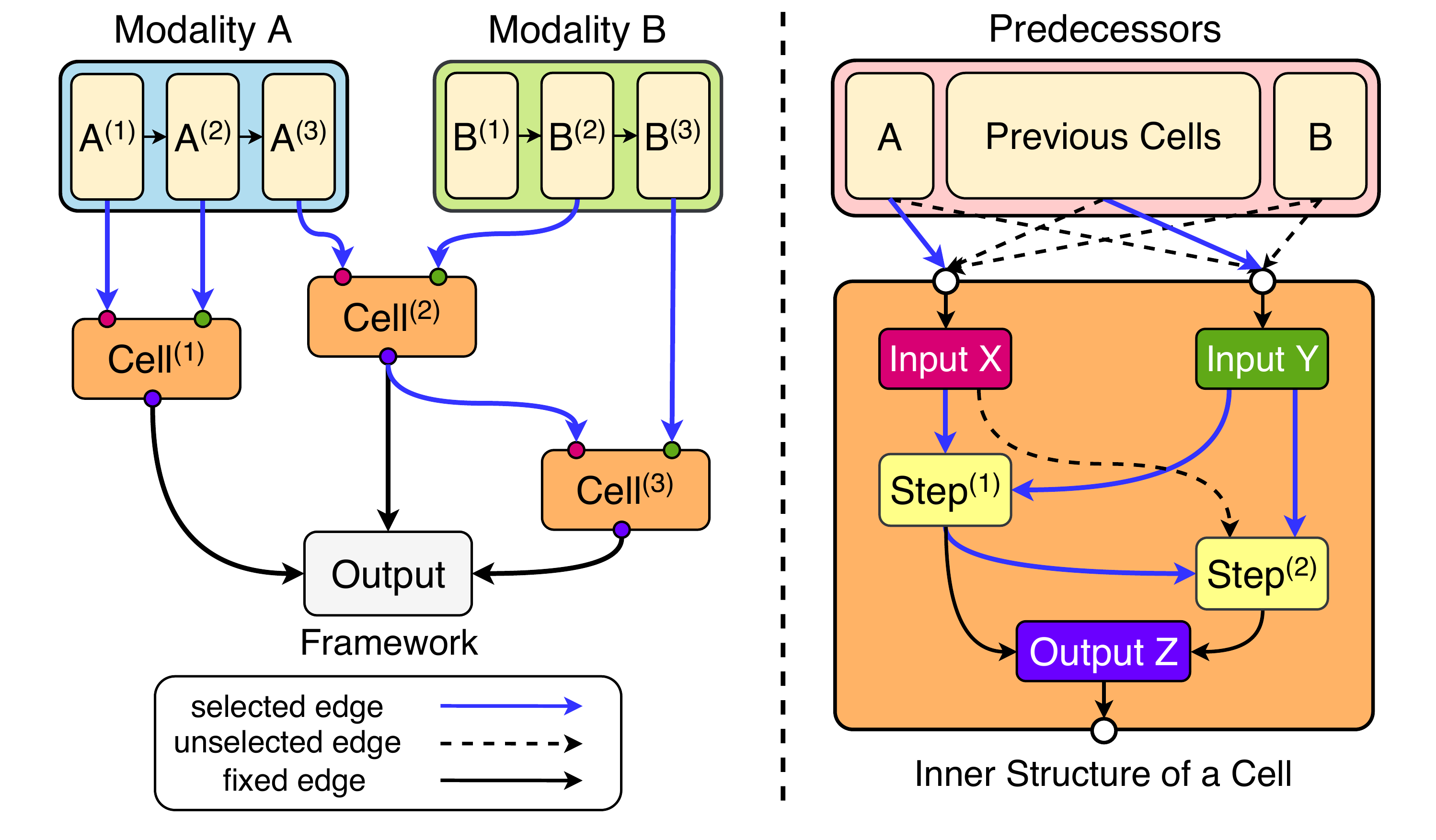}
\end{center}
\vspace{-0.2cm}
\caption{An overview of our BM-NAS framework for multimodal learning. a \emph{Cell} is a searched feature fusion unit that accepts two inputs from modality features or other Cells. In a bilevel fashion, we search the connections between Cells and the inner structures of Cells, simultaneously.}
\label{fig-simple-framework}
\end{figure}

\begin{figure*}
    \begin{center}
    \includegraphics[width=1\textwidth]{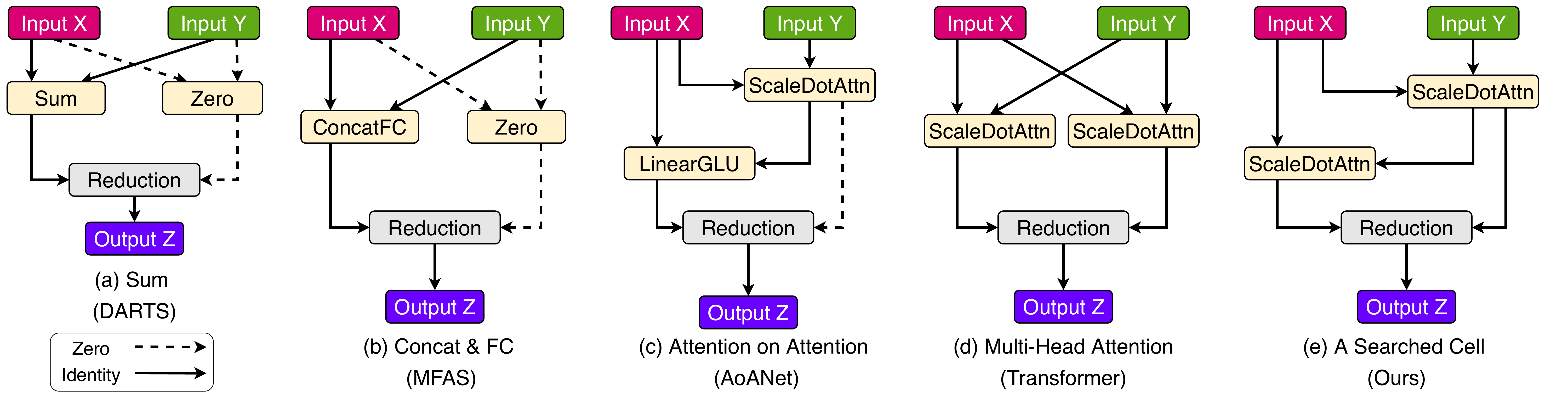}
    \end{center}
    \vspace{-0.3cm}
    \caption{The search space of a Cell in BM-NAS accommodates many existing multimodal fusion strategies. (d) is a two-head version of multi-head attention \cite{vaswani2017attention}, and more heads can be flexibly added by changing the number of inner steps. (e) is the Cell founded by BM-NAS on NTU RGB-D dataset \cite{shahroudy2016ntu}, and it outperforms these existing fusion strategies 
   (see Table \ref{ablation-node-step}). }
    \label{fig-attention-aaai}
\end{figure*}

In this paper, we propose a generalized framework, named Bilevel Multimodal Neural Architecture Search (BM-NAS), to adaptively learn the architectures of DNNs for a variety of multimodal tasks. BM-NAS adopts a \emph{bilevel} searching scheme that it learns the unimodal feature selection strategy at the upper level and the multimodal feature fusion strategy at the lower level, respectively. As shown in the left part of Fig. \ref{fig-simple-framework}, the upper level of BM-NAS consists of a series of feature fusion units, \ie, Cells. The Cells are organized to combine and transform the unimodal features to the task output through a searchable directed acyclic graph (DAG). The right part of Fig. \ref{fig-simple-framework} illustrates the lower level of BM-NAS which learns the inner structures of Cells. A Cell is comprised of several predefined primitive operations. We carefully select the primitive operations such that different combinations of them can form a large variety of fusion modules, as shown in Fig. \ref{fig-attention-aaai}, our search space incorporates benchmark attention mechanisms like multi-head attention (Transformer) \cite{vaswani2017attention} and Attention on Attention (AoA) \cite{huang2019aoa}. The bilevel scheme of BM-NAS is end-to-end learned using the differentiable NAS framework \cite{liu2018darts}. We conduct extensive experiments on three multimodal tasks to evaluate the proposed BM-NAS framework. BM-NAS shows superior performances in comparison with the state-of-the-art multimodal methods. Compared with the existing generalized multimodal NAS frameworks, BM-NAS achieves competitive performances with much less search time and fewer model parameters. To the best of our knowledge, BM-NAS is the first multimodal NAS framework that supports the search of both the unimodal feature selection strategies and the multimodal fusion strategies.

The main contributions of this paper are three-fold.
\begin{enumerate}
    \item Towards a more generalized and flexible design of DNNs for multimodal learning, we propose a new paradigm that employs NAS to search both the unimodal feature selection strategy and the multimodal fusion strategy.
    
    \item We present a novel BM-NAS framework to address the proposed paradigm. BM-NAS makes the architecture of multimodal fusion models fully searchable via a bilevel searching scheme. 
    
    \item We conduct extensive experiments on three multimodal learning tasks to evaluate the proposed BM-NAS framework. Empirical evidences indicate  that both the unimodal feature selection strategy and the multimodal fusion method are significant to the performance of multimodal DNNs.
\end{enumerate}

\section{Related Work}

\subsection{Neural Architecture Search}

Neural architecture search (NAS) aims at automatically finding the optimal neural network architectures for specific learning tasks. NAS can be viewed as a bilevel optimization problem by optimizing the weights and the architecture of DNNs at the same time. Since the network architecture is discrete, traditional NAS methods usually rely on the black-box optimization algorithms, resulting in a extremely large computing cost. For example, searching architectures using reinforcement learning \cite{zoph2016rlnas} or evolution \cite{real2019evolutionnas} would require thousands of GPU-days to find a state-of-the-art architecture on ImageNet dataset \cite{deng2009imagenet} due to low sampling efficiency.

As a result, many methods were proposed for speeding up NAS. From the perspective of engineering, ENAS \cite{pham2018enas} improve the sampling efficiency by weight-sharing. From the perspective of optimization algorithm, PNAS \cite{liu2018pnas} employs sequential model-based optimization (SMBO) \cite{hutter2011smbo}, using surrogate model to predict the performance of an architecture. Monte Carlo tree search (MTCS) \cite{negrinho2017deeparchitect} and Bayesian optimization (BO) \cite{kandasamy2018baysian-nas} are also explored to enhance the sampling efficiency.

Recently, a remarkable efficiency improvement of NAS is achieved by differentiable architecture search (DARTS) \cite{liu2018darts}. DARTS introduces a continuous relaxation of the network architecture, making it possible to search an architecture via gradient-based optimization. However, DARTS only supports the search of unary operations. For specific multimodal tasks, we expect the NAS framework to support the search of multi-input operations, in order to obtain the optimal fusion strategy. In this work, we devise a novel NAS framework named BM-NAS for multimodal learning. BM-NAS follows the optimization scheme of DARTS, however, it novelly introduces a bilevel searching scheme to search the unimodal feature selection strategy and the multimodal fusion strategy simultaneously, enabling an effective search scheme for multimodal fusion.  

\subsection{Multimodal Fusion}

The multimodal fusion techniques for DNNs can be generally classified into two categories: early fusion and late fusion. Early fusion combines low-level features, while late fusion combines prediction-level features. To combine these features, a series of reduction operations such as weighted average \cite{natarajan2012weighted-avg-late-fusion} and bilinear product \cite{teney2018mutan} are proposed in previous works. As each unimodal DNNs backbone could have tens of layers or maybe more, manually sorting out the best intermediate features for multimodal fusion could be exhausting. Therefore, some works propose to enable fusion at multiple intermediate layers. For instance, CentralNet \cite{vielzeuf2018centralnet} and MMTM \cite{joze2020mmtm} join the latent representations at each layer and pass them as auxiliary information for deeper layers. Such methods achieve superior performances on several multimodal tasks including multimodal action recognition \cite{shahroudy2016ntu} and  gesture recognition \cite{zhang2018egogesture}. However, it would largely increase the parameters of multimodal fusion models. 

In recent years, there is an increased interest of introducing the attention mechanisms such as Transformer \cite{vaswani2017attention} to multimodal learning. The multimodal-BERT family \cite{chen2019mmbert-uniter, li2019mmbert-visualbert, lu2019mmbert-vilbert, tan2019mmbert-lxmert} is a typical approach for inter-modal fusion. Moreover, DFAF \cite{gao2019dfaf} shows that intra-modal fusion could also be helpful. DFAF proposes a dynamic attention flow module to mix inter-modal and intra-modal features together through the multi-head attention \cite{vaswani2017attention}. Additional efforts are made to enhance multimodal fusion efficacy of attention mechanisms. For instance, AoANet \cite{huang2019aoa} proposes the attention on attention (AoA) module, showing that adding an attention operation on top of another one could achieve better performance on image captioning task.

Recently, the NAS approaches are making an exciting progress for DNNs, and it shows a huge potential to introduce NAS to multimodal learning. One representative work is MFAS \cite{perez2019mfas}, which employs SMBO algorithm \cite{hutter2011smbo} to search multimodal fusion strategies given the unimodal backbones. But as SMBO is a black-box optimization algorithm, every update step requires a bunch of DNNs to be trained, leading to the inefficiency of MFAS. Besides, MFAS only use concatenation and fully connected (FC) layers for unimodal feature fusion, and the stack of FC layers would be a heavy burden for computing. Further work like MMIF \cite{peng2020mmif} and  3D-CDC \cite{yu20213dcdc} adopt the efficient DARTS algorithm \cite{liu2018darts} for architecture search but only support the search of unary operations on graph edges and use summation on every intermediate node for reduction. MMnas \cite{yu2020mmnas} allows searching the attention operations but the topological structure of the network is fixed during architecture search. 

Different from these related works, our proposed BM-NAS supports to search both the unimodal feature selection strategy and the fusion strategy of multimodal DNNs. BM-NAS introduces a bilevel searching scheme. The upper level of BM-NAS supports both intra-modal and inter-modal feature selection. The lower level of BM-NAS searches the fusion operations within every intermediate step. Each step can flexibly form the summation, concatenation, multi-head attention \cite{vaswani2017attention}, attention on attention \cite{huang2019aoa}, or any other unexplored fusion mechanisms. BM-NAS is a generalized and efficient NAS framework for multimodal learning. In experiments we show that BM-NAS can be applied to various multimodal tasks regardless of the modalities or backbone models.

\section{Methodology}

\begin{figure*}
    \begin{center}
    \includegraphics[width=1\textwidth]{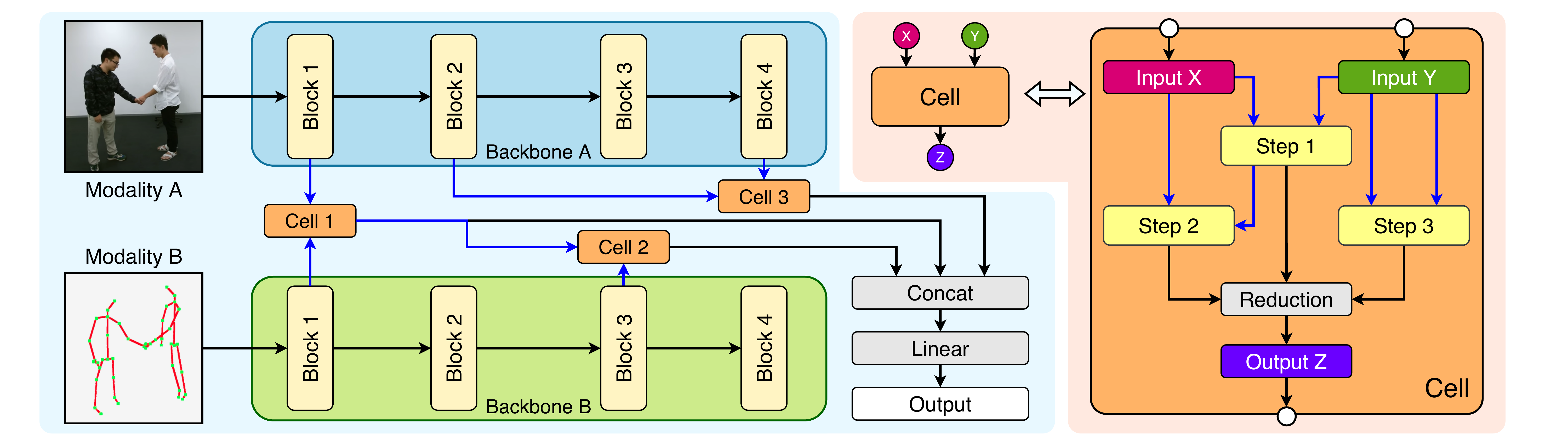}
    \end{center}
    \vspace{-0.2cm}
    \caption{An example of a multimodal fusion network found by BM-NAS, which consists of a bilevel searching scheme, we denote searched edges in blue, and fixed edges in black. \textbf{Left}: The upper level BM-NAS. The input features are extracted by pretrained unimodal models. Each \emph{Cell} accepts two inputs from its predecessors, \ie, any unimodal feature or previous Cell. \textbf{Right}: The lower level BM-NAS. Within a Cell, each \emph{Step} denotes a primitive operation selected from a predefined operation pool. The topologies of Cells and Steps are both searchable. The numbers of Cells and Steps are hyper-parameters such that BM-NAS can be adapted to a variety of multimodal tasks with different scales.}
    \label{framework}
\vspace{-0.2cm}
\end{figure*}



In this work, we propose a generalized NAS framework, named Bilevel Multimodal NAS (BM-NAS), to search the architectures of multimodal fusion DNNs. More specifically, BM-NAS searches a Cell-by-Cell architecture in a bilevel fashion. The upper level architecture is a directed acyclic graph (DAG) of the input features and Cells. The lower level architecture is a DAG of inner step nodes within a Cell. Each inner step node is a bivariate operation drawn from a predefined pool. The bilevel searching scheme ensures that BM-NAS can be easily adapted to various multimodal learning tasks regardless of the types of modalities. In the following, we discuss the unimodal feature extraction in Section \ref{sec:unimodal}, the upper and lower levels of BM-NAS in Sections \ref{sec:upper} and \ref{sec:lower}, along with the architecture search algorithm and evaluation in Section \ref{sec:algorithm}.


\subsection{Unimodal feature extraction}
\label{sec:unimodal}
By following previous multimodal fusion works, such as CentralNet \cite{vielzeuf2018centralnet}, MFAS \cite{perez2019mfas} and MMTM \cite{joze2020mmtm}, we also employ the pretrained unimodal backbone models as the feature extractors. We use the outputs of their intermediate layers as raw features (or intermediate blocks if the model has a block-by-block structure like ResNeXt \cite{xie2017resnext}). 

Since the raw features vary in shapes, we reshape them by applying pooling, interpolation, and fully connected layers on spatial, temporal, and channel dimensions, successively. By doing so, we reshape all the raw features to the shape of $(N, C, L)$, such that we can easily perform fusion operations between  features of different modalities. Here $N$ is the batch size, $C$ is the embedding dimension or the number of channels, $L$ is the sequence length. 

\subsection{Upper Level: Cells for Feature Selection}
\label{sec:upper}

The upper level of BM-NAS searches the unimodal feature selection strategy and it consists of a group of Cells. Formally, suppose we have two modalities A and B, and two pretrained unimodal models for each modality. Let $\{A^{(i)}\}$ and $\{B^{(i)}\}$ indicate the modality features extracted by the backbone models. We formulate the upper level nodes in an ordered sequence $\mathcal{S}$, as
\begin{equation}
\mathcal{S} =[ A^{(1)}, ..., A^{(N_A)},B^{(1)},...,B^{(N_B)},\text{Cell}^{(1)},..., \text{Cell}^{(N)} ]. \notag
\end{equation}
Under the setting of $\mathcal{S}$, both inter-modal fusion and intra-modal fusion are considered in BM-NAS.

\noindent\textbf{Feature selection.}
By adopting the continuous relaxation in differentiable architecture search scheme \cite{liu2018darts}, all predecessors of $\text{Cell}^{(i)}$ will be connected to $\text{Cell}^{(i)}$ through weighted edges at the searching stage. This directed complete graph between Cells is called the \emph{hypernet}. For two upper level nodes $s^{(i)}, s^{(j)} \in \mathcal{S}$, let $\alpha^{(i,j)}$ denote the edge weight between $s^{(i)}$ and $s^{(j)}$. Each edge is a unary operation $g$ selected from a function set $\mathcal{G}$ including

(1) $\mathrm{Identity}(x) = x$, \ie, selecting an edge.

(2) $\mathrm{Zero}(x) = 0$, \ie, discarding an edge.

Then, the mixed edge operation $\overline{g}^{(i,j)}$ on edge $(i,j)$ is
\begin{equation}
\overline{g}^{(i,j)}(s) = \sum\limits_{g \in \mathcal{G}} \frac{\text{exp}(\alpha_g^{(i,j)})} {\sum\limits_{g' \in \mathcal{G}} \text{exp}( \alpha_{g'}^{(i,j)} ) } g(s) .
\end{equation}
A Cell $s^{(j)}$ receives inputs from all its predecessors, as
\begin{equation}
s^{(j)} = \sum\limits_{i<j}\overline{g}^{(i,j)}(s^{(i)}) .
\end{equation}


In evaluation stage, the network architecture is discretized that an input pair $(s^{(i)}, s^{(j)})$\footnote{We enforce the Cells to have different predecessors.} will be selected for $s^{(k)}$ if
\begin{equation}
(i,j) = \argmax_{i<j<k,\ g\in \mathcal{G}}(\alpha_g^{(i,k)} \cdot \alpha_g^{(j,k)}) .
\end{equation}
It is worth noting that, compared with searching the feature pairs directly, the Cell-by-Cell structure significantly reduces the complexity of the search space for unimodal feature selection. For an input pair from two feature sequences $[ A^{(1)}, ..., A^{(N_A)}]$ and $[B^{(1)},...,B^{(N_B)}]$, the number of candidate choices is $2(N_A+N_B)$ under the Cell-by-Cell search setting. It is much smaller than $C_{N_A+N_B}^2$, the number of candidates under the pairwise search setting. 
%



\subsection{Lower Level: Multimodal Fusion Strategy}
\label{sec:lower}

The lower level of BM-NAS searches the multimodal fusion strategy, \ie, the inner structure of Cells. Specifically, a Cell is a DAG consisting of a set of inner step nodes. The inner step nodes are the primitive operations drawn from a predefined operation pool. We introduce our predefined operation pool in the following.


\noindent\textbf{Primitive operations.}
All the primitive operations take two tensor inputs $x, y$, and outputs a tensor $z$, where $x,y,z \in \mathbb{R}^{N\times C\times L}$. 

(1) $\mathrm{Zero}(x, y)$: The $\mathrm{Zero}$ operation discards an inner step completely. It will be helpful when BM-NAS decides to use only a part of the inner steps.
\begin{equation}
\mathrm{Zero}(x, y) = 0 .
\end{equation}

(2) $\mathrm{Sum}(x, y)$: The DARTS \cite{liu2018darts} framework uses summation to combine two features as
\begin{equation}
\mathrm{Sum}(x, y) = x + y .
\end{equation}

(3) $\mathrm{Attention}(x, y)$: We use the scaled dot-product attention \cite{vaswani2017attention}. As a standard attention module usually takes three inputs namely query, key, and value, we let the query be $x$, the key and value be $y$, which is also known as the guided-attention \cite{yu2020mmnas}.
\begin{equation}
\mathrm{Attention}(x, y) = \mathrm{Softmax}(\frac{x y^T}{\sqrt{C}}y).
\end{equation}

(4) $\mathrm{LinearGLU}(x, y)$: A linear layer with the gated linear unit (GLU) \cite{dauphin2017glu}. Let $W_1, W_2 \in \mathbb{R}^{C \times C}$ and $\odot$ be element-wise multiplication, then $\mathrm{LinearGLU}$ is
\begin{align}
\mathrm{LinearGLU}(x, y) &= \mathrm{GLU}(x W_1, y W_2) \\
&= x W_1 \odot \mathrm{Sigmoid}(y W_2) . \notag
\end{align}

(5) $\mathrm{ConcatFC}(x, y)$: $\mathrm{ConcatFC}$ stands for passing the concatenation of $(x, y)$ to a fully connected (FC) layer with ReLU activation \cite{nair2010relu}. The FC layer reduces the channel numbers from $2C$ to $C$. Let $W \in \mathbb{R}^{2C \times C}, b \in \mathbb{R}^{C}$, then $\mathrm{ConcatFC}$ is
\begin{equation}
\mathrm{ConcatFC}(x, y) = \mathrm{ReLU} (\mathrm{Concat}(x, y) W + b) .
\end{equation}

We elaborately choose these primitive operations such that they can be flexibly combined to form various feature fusion modules. In Fig. \ref{fig-attention-aaai}, we show that the search space of lower level BM-NAS accommodates many benchmark multimodal fusion strategies such as the summation used in DARTS \cite{liu2018darts}, the ConcatFC used in MFAS \cite{perez2019mfas}, the multi-head attention used in Transformer \cite{vaswani2017attention}, and the Attention on Attention used in AoANet \cite{huang2019aoa}. There also remains flexibility to discover other better fusion modules for specific multimodal learning tasks.

\noindent\textbf{Fusion strategy.}
In searching stage, the inner step set of $\text{Cell}^{(n)}$ is an ordered feature sequence $\mathcal{T}_n$, 
\begin{equation}
\mathcal{T}_n = [x, y, \text{Step}^{(1)}, ...,\text{Step}^{(M)}] .
\end{equation}

An inner step node $t^{(i)}$ transforms two input nodes $t^{(j)}, t^{(k)}$ to its output through an average over the primitive operation pool $\mathcal{F}$, as
\begin{equation}
\overline{f}^{(i)}(t^{(j)}, t^{(k)}) = \sum\limits_{f \in \mathcal{F}} \frac{\text{exp}(\gamma_{f}^{(i)})} {\sum\limits_{f' \in \mathcal{F}} \text{exp}( \gamma_{f'}^{(i)} ) } f(t^{(j)}, t^{(k)}),
\end{equation}
where $\gamma$ is the weights of primitive operations. In the evaluation stage, the optimal operation of an inner step node is derived as,
\begin{equation}
f^{(i)} = \argmax_{f\in \mathcal{F}} \gamma_{f}^{(i)} .
\end{equation}




The continuous relaxation of the edges with weights $\beta$ between inner step nodes is similar to the upper level. For a simplicity, we omit the formulation in this paper. Note that unlike the upper level BM-NAS, the pairwise inputs in a Cell can be chosen repeatedly\footnote{We don't enforce the step nodes to have different predecessors.}, so the inner steps can form structures like multi-head attention \cite{vaswani2017attention}. 

\subsection{Architecture Search and Evaluation}
\label{sec:algorithm}
\begin{center}
    \begin{algorithm}[t]
        \caption{Bilevel Multimodal NAS (BM-NAS)}
        \SetAlgoLined
        \KwResult{The genotype of fusion networks.}
        Initialize architecture parameters $\alpha, \beta, \gamma$ and model parameters $w$\;
        Initialize \emph{genotype} based on $\alpha, \beta, \gamma$, set \emph{genotype\_best} = \emph{genotype}\;
        Construct \emph{hypernet} based on \emph{genotype\_best}\;
         \While{$\mathcal{L}$ not converged}{
          Update $\omega$ on training set\;
          Update ($\alpha$, $\beta$, $\gamma$) on validation set\;
          Derive upper level \emph{genotype} based on $\alpha$, derive lower level \emph{genotype} based on $\beta, \gamma$\;
          Update \emph{hypernet} based on \emph{genotype}\;
          \If{higher validation accuracy is reached}{
            Update \emph{genotype\_best} using \emph{genotype}\; 
          }
         }
        Return \emph{genotype\_best}\;
        \label{algorithm1}
    \end{algorithm}
\vspace{-0.4cm}

\end{center}

\begin{figure}[tbp]
    \begin{center}
    \includegraphics[width=0.9\linewidth]{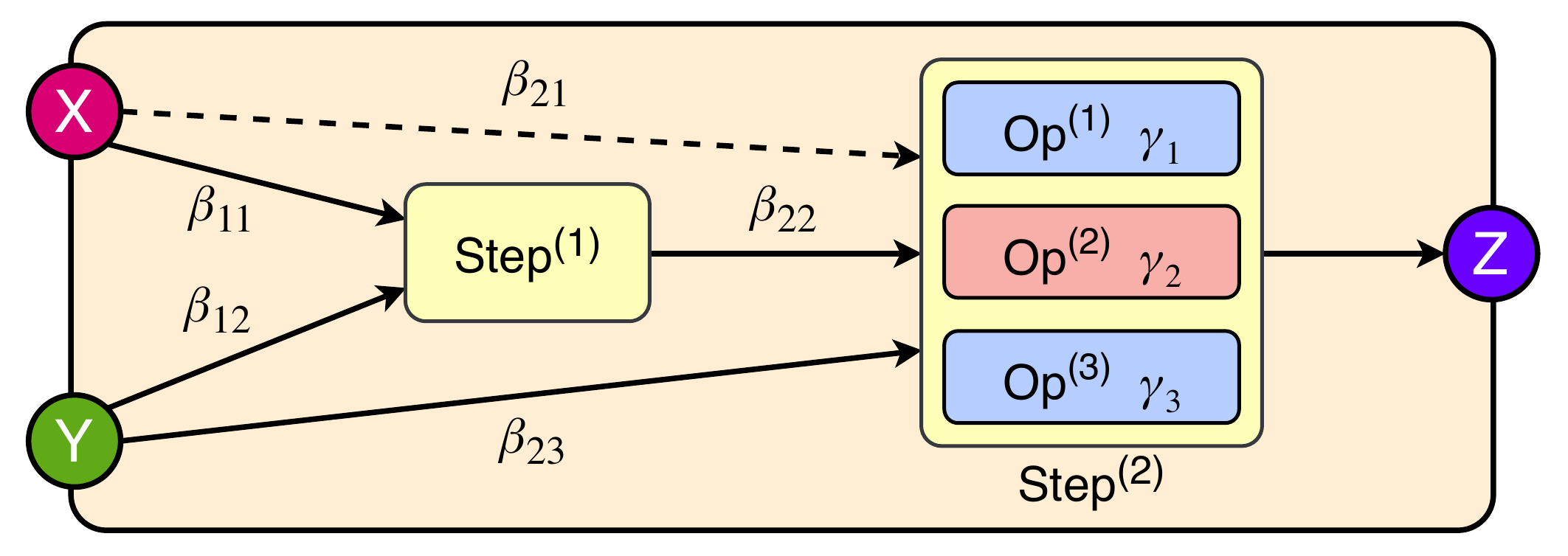}
    \end{center}
    \vspace{-0.4cm}
    \caption{Architecture parameters $\beta$ and $\gamma$ of a \emph{Cell}.}
    \label{fig-beta-gamma}
    \vspace{-0.2cm}
\end{figure}

\noindent\textbf{Architecture Parameters.} The function of the weights of primitive operations ($\beta$) and inner step nodes edges ($\gamma$) is shown in Fig. \ref{fig-beta-gamma}, $\beta$ is used for feature selection within the cell, selecting two inputs for each inner step node. And $\gamma$ is used for operation selection on each inner step node.

\noindent\textbf{Search algorithm.} In Sections \ref{sec:upper} and \ref{sec:lower}, we introduced three variable $\alpha, \beta, \gamma$ as the architecture parameters. Algorithm \ref{algorithm1} shows the searching process of BM-NAS, which follows DARTS \cite{liu2018darts} to optimize $\alpha, \beta, \gamma$ and model weights $w$, alternatively. In Algorithm \ref{algorithm1}, the model in searching stage is called \emph{hypernet} since all the edges and nodes are mixed operations. The searched structure description of the fusion network is called \emph{genotype}. 

\noindent\textbf{Implementation details.} In order to make the whole BM-NAS framework searchable and flexible, Cells/inner step nodes should have the same number of inputs and output, so they can be put together in arbitrary topological order. The two-input setting follows the benchmark NAS frameworks (DARTS \cite{liu2018darts}), MFAS \cite{perez2019mfas}, MMIF \cite{peng2020mmif}, \etc). They all have only two searchable inputs for each Cell/step node. Also, it requires no extra effort to let the Cells or step nodes support 3 or more inputs, by just adding ternary (or other arbitrary) operations into the primitive operation pool. 

\noindent\textbf{Evaluation.} In architecture evaluation, we select the \emph{genotype} with the best validation performance as the searched fusion network. Then we combine the training and validation sets together to train the unimodal backbones and the searched fusion network jointly.

\section{Experiments}

In this work we evaluate the BM-NAS on three multimodal tasks, including (1) the multi-label movie genre classification task on MM-IMDB dataset \cite{arevalo2017gmu}, (2) the multimodal action recognition task on NTU RGB-D dataset \cite{shahroudy2016ntu}, and (3) the multimodal gesture recognition task on EgoGesture dataset \cite{zhang2018egogesture}. In the following, we discuss the experiments on the three tasks in Sections \ref{sec:mm}, \ref{sec:ntu}, and \ref{sec:ego}, respectively. We perform computing efficiency analysis in Section \ref{sec:time}. We further evaluate the search strategies of the proposed BM-NAS framework in Sections \ref{sec:ablation}.

In addition, we present the examples of these tasks, thorough discussion of hyper-parameter configurations, visualization of searched architectures and their performances during the searching stage (hypernets) and evaluation stage (final model) in our supplementary material.


\subsection{MM-IMDB Dataset}
\label{sec:mm}
MM-IMDB dataset \cite{arevalo2017gmu} is a multi-modal dataset collected from the Internet Movie Database, containing posters, plots, genres and other meta information of 25,959 movies. We conduct multi-label genre classification on MM-IMDB using posters (RGB images) and plots (text) as the input modalities. There are 27 non-mutually exclusive genres in total, including \textit{Drama, Comedy, Romance}, etc. Since the number of samples in each class is highly imbalanced, we only use 23 genres for classification. The classes of \textit{News, Adult, Talk-Show, Reality-TV} are discarded since they only count for 0.10\% in total. We adopt the original split of the dataset where 15,552 movies are used for training, 2,608 for validation and 7,799 for testing.

For a fair comparison with other explicit multimodal fusion methods, we use the same backbone models. Specifically, we use Maxout MLP \cite{goodfellow2013maxout} as the backbone of text modality and VGG Transfer \cite{simonyan2014vgg} as the backbone of RGB image modality. For BM-NAS, we adopt a setting of 2 fusion Cells and 1 step/Cell. For inner step representations, we set $C=192,L=16$.

Table \ref{mmimdb-exp} shows that BM-NAS achieves the best Weighted F1 score in comparison with the existing multimodal fusion methods. Notice that as the class distribution of MM-IMDB is highly imbalanced, Weighted F1 score is in fact a more reliable metric for measuring the performance of multi-label classification than other kinds of F1 score.

\subsection{NTU RGB-D Dataset}


\begin{table}[tbp]
\small
\begin{minipage}{1.0\linewidth}
\centering
\caption{Multi-label genre classification results on MM-IMDB dataset. Weighted F1 (F1-W) is reported.}
\begin{tabular}{c|c|c}
      \hline
      Method & Modality & F1-W(\%) \\
      \hline
      \multicolumn{3}{c}{Unimodal Methods} \\
      \hline
      Maxout MLP (ICML13) & Text & 57.54 \\
      VGG Transfer (ICLR15) & Image & 49.21 \\
      \hline
      \multicolumn{3}{c}{Multimodal Methods} \\
      \hline
      Two-stream (NIPS14) & Image + Text & 60.81\\
      GMU (ICLR17) & Image + Text & 61.70 \\
      CentralNet (ECCV18) & Image + Text & 62.23\\
      MFAS (CVPR19) & Image + Text & 62.50 \\
      BM-NAS (ours) & Image + Text & \bf{62.92 $\pm$ 0.03} \\
      \hline
\end{tabular}  
\label{mmimdb-exp}
\end{minipage}
\vspace{-0.3cm}
\end{table}


\begin{table}[tbp]
    \small
	\begin{minipage}{1.0\linewidth}
	\centering
	\caption{Action recognition results on NTU RGB-D dataset. }
	\begin{tabular}{c|c|c}
	    \hline
        Method & Modality & Acc(\%) \\
        \hline
        \multicolumn{3}{c}{Unimodal Methods} \\
        \hline
        Inflated ResNet-50 (CVPR18) & Video & 83.91 \\
        Co-occurrence (IJCAI18) & Pose & 85.24 \\
        \hline
        \multicolumn{3}{c}{Multimodal Methods} \\
        \hline
        Two-stream (NIPS14) & Video + Pose & 88.60 \\
        GMU (ICLR17) & Video + Pose & 85.80 \\
        MMTM (CVPR20) & Video + Pose & 88.92 \\
        CentralNet (ECCV18) & Video + Pose & 89.36 \\
        MFAS (CVPR19) & Video + Pose & 89.50 $\pm$ 0.60 \\
        BM-NAS (ours)  & Video + Pose & \bf{90.48 $\pm$ 0.24} \\
        \hline
	\end{tabular}  
	\label{ntu-exp}
	\end{minipage}
\end{table}

\label{sec:ntu}
The NTU RGB-D dataset \cite{shahroudy2016ntu} is a large scale multimodal action recognition dataset, containing a total of 56,880 samples with 40 subjects, 80 view points, and 60 classes of daily activities. In this work we use the skeleton and RGB video modality for fusion experiments. We measure the performance of methods using cross-subject (CS) accuracy. We follow the dataset split of MFAS \cite{perez2019mfas}. In detail, we use subjects 1, 4, 8, 13, 15, 17, 19 for training, 2, 5, 9, 14 for validation, and the rest for test. There are 23760, 2519 and 16558 samples in the training, validation, and test dataset, respectively.

For a fair comparison, we use two CNN models, the Inflated ResNet-50 \cite{baradel2018inflated} for video modality and Co-occurrence \cite{li2018co} for skeleton modality as backbones, ensuring all the methods in our experiments share the same backbones. We test the performances of MFAS \cite{perez2019mfas}, MMTM \cite{joze2020mmtm}, and the proposed BM-NAS using our data prepossessing pipeline, such that the performances of these methods are not the same as they were original reported. For BM-NAS, we use 2 fusion Cells and 2 Steps/Cell. For inner step representations we set $C=128,L=8$. 

In Table \ref{ntu-exp}, our method achieves an cross-subject accuracy of $90.48\%$, showing an state-of-the-art result on NTU RGB-D \cite{shahroudy2016ntu} with video and pose modalities.

\begin{figure}[tb]
  \begin{center}
  \includegraphics[width=0.8\linewidth]{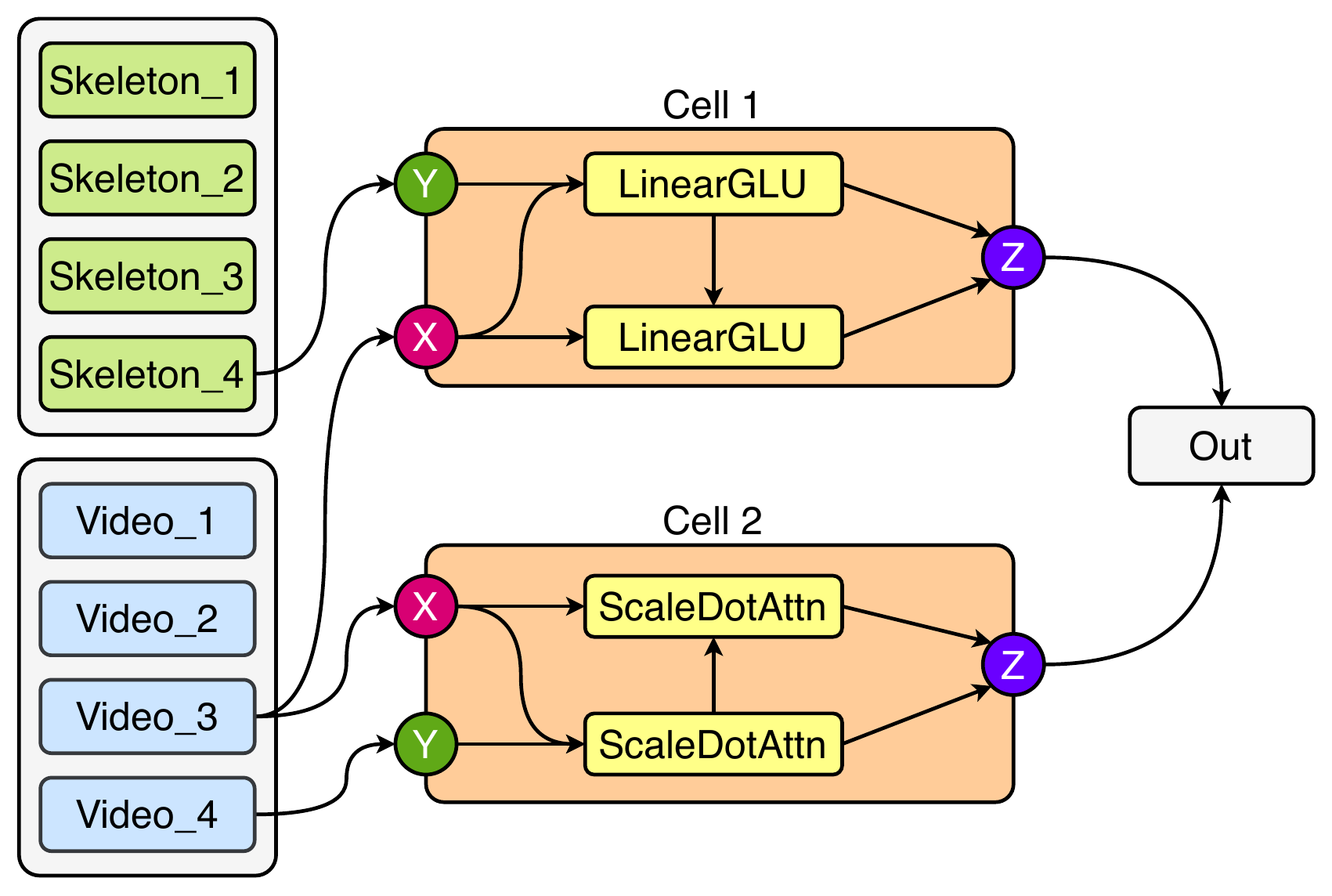}
  \end{center}
  \vspace{-0.4cm}
  \caption{Best model found on NTU RGB-D dataset.}
  \vspace{-0.4cm}
  \label{fig-best-ntu-model}
\end{figure}


\subsection{EgoGesture Dataset}
\label{sec:ego}
The EgoGesture dataset \cite{zhang2018egogesture} is a large scale multimodal gesture recognition dataset, containing 24,161 gesture samples of 83 classes collected from 50 distinct subjects and 6 different scenes. We follow the original cross-subject split of EgoGesture dataset \cite{zhang2018egogesture}. There are 14,416 samples for training, 4,768 for validation, and 4,977 for testing.

We use the ResNeXt-101 \cite{kopuklu2019real} as the backbone on both RGB and depth video modality. As former works like CentralNet \cite{vielzeuf2018centralnet} and MFAS \cite{perez2019mfas} did not perform experiments on this dataset, we compared our method with other unimodal and multimodal methods, especially MMTM \cite{joze2020mmtm}, MTUT \cite{gupta2019mtut} and 3D-CDC \cite{yu20213dcdc}. Since we do not search for the backbone, we compared with 3D-CDC-NAS2, which 
also uses ResNeXt-101 as the backbones. For our BM-NAS, we use 2 fusion Cells and 3 steps/Cell, for inner step representations we set $C=128, L=8$.

Table \ref{tab:param-list} reports the experiment results on EgoGesture \cite{zhang2018egogesture}. Comparing to 3D-CDC, which requires 3 groups of backbone models trained under different video frame rates (8, 16 and 32 FPS), our BM-NAS only requires the 32 FPS ones, and is generalized to all kinds of modalities. In general, BM-NAS achieves a state-of-the-art fusion performance, showing that BM-NAS is effective for enhancing gesture recognition performance on EgoGesture dataset.

\begin{table}[tbp]
    \small
	\centering
	\caption{Gesture recognition results on EgoGesture dataset. We use ResNext-101 as backbones for both RGB and depth modality for our BM-NAS method.}

    \resizebox{\linewidth}{!}{

	\begin{tabular}{c|c|c}
        \hline
	    Method         & Modality    & Acc(\%) \\
        \hline
        \multicolumn{3}{c}{Unimodal Methods} \\
        \hline
        VGG-16 + LSTM (NIPS14) & RGB         & 74.70    \\
        C3D + LSTM + RSTTM (ICCV15) & RGB         & 89.30    \\
        I3D (CVPR17)           & RGB         & 90.33    \\
        ResNext-101 (FG19)    & RGB         & 93.75    \\
        VGG-16 + LSTM (CVPR14)   & Depth       & 77.70    \\
        C3D + LSTM + RSTTM (CVPR16) & Depth       & 90.60    \\
        I3D (CVPR17)             & Depth         & 89.47    \\
        ResNeXt-101 (FG19)   & Depth       & 94.03    \\
        \hline
        \multicolumn{3}{c}{Multimodal Methods} \\
        \hline
        VGG-16 + LSTM (CVPR17)   & RGB + Depth & 81.40    \\
        C3D + LSTM + RSTTM (CVPR19)  & RGB + Depth & 92.20    \\
        I3D (CVPR17)      & RGB + Depth & 92.78    \\
        MMTM (CVPR20)  & RGB + Depth & 93.51    \\
        MTUT (3DV19) & RGB + Depth & 93.87    \\
        3D-CDC-NAS2 (TIP21) & RGB + Depth & 94.38 \\
        BM-NAS (ours)           & RGB + Depth & \bf{94.96 $\pm$ 0.07}    \\
        \hline
	\end{tabular}  

    }
	\label{tab:param-list}
    \vspace{-0.2cm}
\end{table}

\begin{table}[tbp]
    \small
	\begin{minipage}{1.0\linewidth}
	\centering
	\caption{Model size and performance on NTU RGB-D. }
	\begin{tabular}{c|c|c|c}
	    \hline
        Method & Dataset & Parameters & Acc(\%) \\
        \hline
        MMTM (CVPR20) & NTU & 8.61 M & 88.92 \\
        MFAS (CVPR19) & NTU & 2.16 M & 89.50 \\
        BM-NAS (ours) & NTU & \bf{0.98 M} & \bf{90.48} \\
        \hline
	\end{tabular}  
	\label{model-params}
	\end{minipage}
    \small
    \begin{minipage}{1.0\linewidth}
        \centering
        \vspace{0.2cm}
        \caption{Search cost (GPU$\cdot$hours) of generalized multimodal NAS methods.}
        \begin{tabular}{c|c|c}
            \hline
            Method & MM-IMDB & NTU \\
            \hline
            MFAS (CVPR19) & 9.24 &  603.64 \\
            BM-NAS (ours) & \bf{0.89} &  \bf{38.6} \\
            \hline
        \end{tabular}  
        \label{search-time}
        \end{minipage}
\vspace{-0.3cm}
\end{table}


\subsection{Computing Efficiency}
\label{sec:time}

\noindent\textbf{Model size.}
Table \ref{model-params} compares the model sizes of different multimodal fusion methods on NTU RGB-D \cite{shahroudy2016ntu}. All three methods share exactly the same unimodal backbones. Compared with the manually designed fusion model MMTM \cite{joze2020mmtm} and the fusion model searched by MFAS \cite{perez2019mfas}, our BM-NAS achieves better performance with fewer model parameters.

\noindent\textbf{Search cost.}
Table \ref{search-time} compares the search cost of generalized multimodal NAS frameworks including MFAS and our BM-NAS. Thanks to the efficient differentiable architecture search framework \cite{liu2018darts}, BM-NAS is at least 10x faster than MFAS when searching on MM-IMDB \cite{arevalo2017gmu} and NTU RGB-D.

\begin{table}[tb]
    \small
	\begin{minipage}{1.0\linewidth}
	\centering
	\caption{Ablation study for feature selection.}
	\begin{tabular}{c|c|c}
	    \hline
        Features            & Dataset   & Accuracy(\%) \\
        \hline
        Random              & NTU       & 86.35 $\pm$ 0.68 \\
        Late fusion         & NTU       & 89.49 $\pm$ 0.15 \\ 
        \hline
        Searched (MFAS)     & NTU       & 89.50 $\pm$ 0.60\\ 
        Searched (BM-NAS)   & NTU       & \bf{90.48} $\pm$ 0.24 \\
        \hline
	\end{tabular}  
	\label{ablation-feature-selection}
	\end{minipage}
\end{table}

\subsection{Ablation Study}
\label{sec:ablation}
In this section, we conduct ablation study to verify the effectiveness of the unimodal feature selection strategy and the multimodal fusion strategy, respectively.

\noindent\textbf{Unimodal feature selection.}
Table \ref{ablation-feature-selection} compares different unimodal feature selection strategies on NTU RGB-D. We compare the best strategy found by BM-NAS against random selection, late fusion, and the best strategy found by MFAS. For all the random baselines, the inner structure of Cells are the same. We randomly selects the input features and the connections between Cells, and report the result averaged over 5 trials. For the late fusion baseline, we concatenate feature pair (\emph{Video\_4},\emph{Skeleton\_4}) in Fig. \ref{fig-best-ntu-model}. MFAS selects four feature pairs: (\emph{Video\_4}, \emph{Skeleton\_4}), (\emph{Video\_2}, \emph{Skeleton\_4}), (\emph{Video\_2}, \emph{Skeleton\_2}), and (\emph{Video\_4}, \emph{Skeleton\_4}). As shown in Table \ref{ablation-feature-selection}, the searched feature selection strategy is better than all baselines, demonstrating that a better unimodal feature selection strategy benefits the multimodal fusion performance. As shown in Table \ref{ablation-feature-selection}, the searched feature selection strategy is better than all baselines, demonstrating that a better unimodal feature selection strategy benefits the multimodal fusion performance.



\begin{table}[t]
    \small
	\begin{minipage}{1.0\linewidth}
	\centering
	\caption{Ablation study for fusion strategy.}
	\begin{tabular}{c|c|c|c}
	    \hline
        Fusion      & Framework                 & Dataset   & Acc (\%) \\
        \hline
        Sum         & DARTS (ICLR19)         & NTU       &87.64 \\
        ConcatFC    & MFAS (CVPR19)          & NTU       &89.20 \\
        MHA         & Transformer (NIPS17)   & NTU       &88.29 \\
        AoA         & AoANet (ICCV19)        & NTU       &89.11 \\
        Searched    & BM-NAS                    & NTU       &\bf{90.48} \\
        \hline
	\end{tabular}  
	\label{ablation-node-step}
	\end{minipage}
\vspace{-0.2cm}
\end{table}

\noindent\textbf{Multimodal fusion strategy.}
Table \ref{ablation-node-step} evaluates different multimodal fusion strategies on NTU RGB-D. All the strategies in Table \ref{ablation-node-step} adopt the same feature selection strategy. We compare the best Cell structure found by BM-NAS against the summation used in DARTS \cite{liu2018darts}, the ConcatFC used in MFAS \cite{perez2019mfas}, the multi-head attention (MHA) used in Transformer \cite{vaswani2017attention}, and the attention on attention (AoA) used in AoANet \cite{huang2019aoa}. All these fusion strategies can be formed as certain combinations of our predefined primitive operations, as shown in Fig. \ref{fig-attention-aaai}. In Table \ref{ablation-node-step}, the fusion strategy derived by BM-NAS outperforms the baseline strategies, showing the effectiveness of searching fusion strategy for multimodal fusion models.


\section{Conclusion}
In this paper, we have presented a novel multimodal NAS framework BM-NAS to learn the architectures of multimodal fusion models via a bilevel searching scheme. To our best knowledge, BM-NAS is the first NAS framework that supports to search both the unimodal feature selection and the multimodal fusion strategies for multimodal DNNs. In experiments, we have demonstrated the effectiveness and efficiency of BM-NAS on various multimodal learning tasks.

\appendix

\section{Datasets and Tasks}
\begin{figure}[h]
    \begin{center}
    \includegraphics[width=1\linewidth]{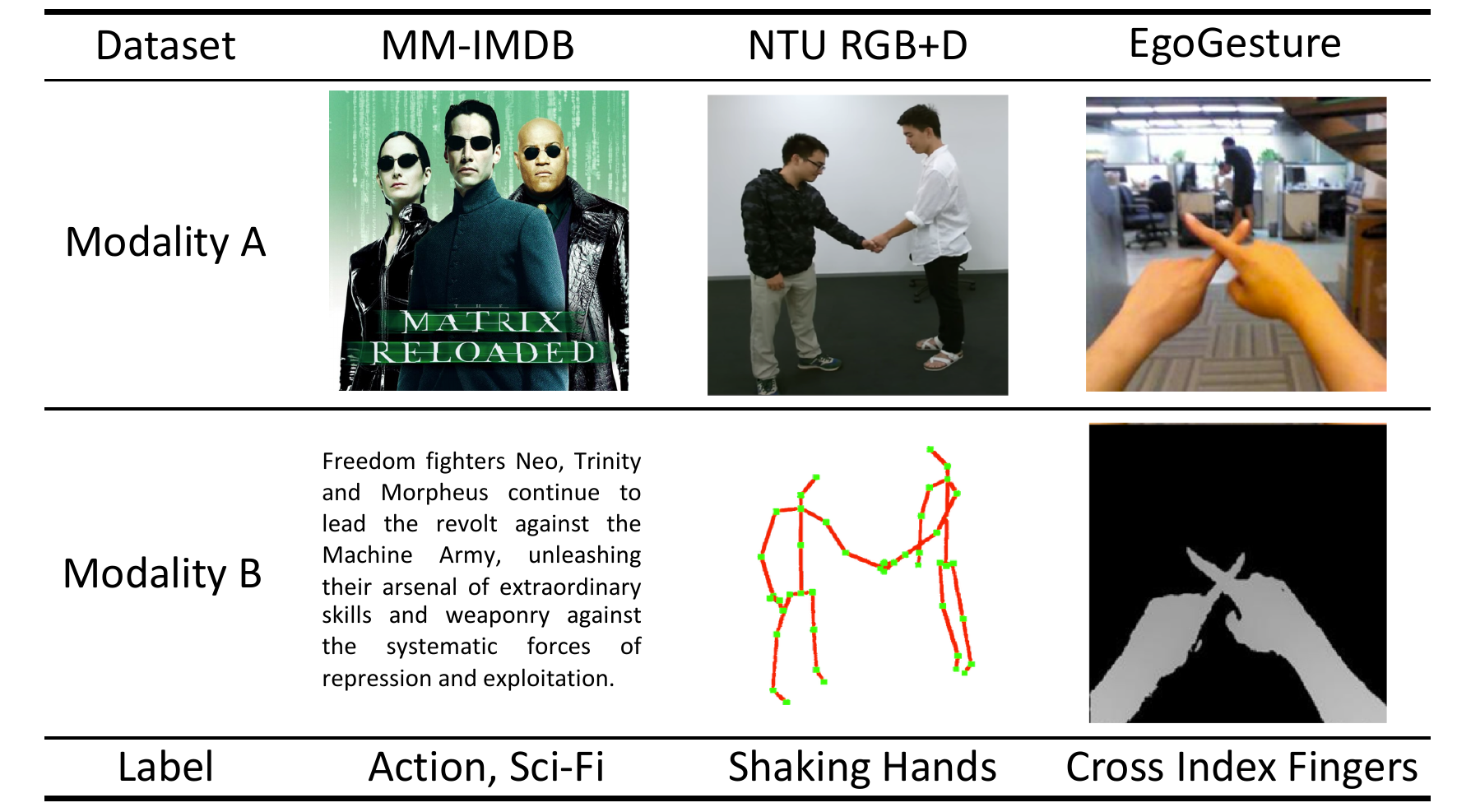}
    \end{center}
    \caption{Examples of the evaluation datasets.}
    \label{fig-dataset-demo}
\end{figure}

In this work we evaluate the BM-NAS on three multimodal tasks, including (1) the multi-label movie genre classification task on MM-IMDB dataset \cite{arevalo2017gmu}, (2) the multimodal action recognition task on NTU RGB-D dataset \cite{shahroudy2016ntu}, and (3) the multimodal gesture recognition task on EgoGesture dataset \cite{zhang2018egogesture}. Examples of these tasks are shown in Fig. \ref{fig-dataset-demo}.

\begin{table*}[b]
    \begin{minipage}[b]{1\textwidth}
    \setlength\tabcolsep{3.5pt} 
    \begin{tabular}{|c|c|c|c|c|c|c|c|c|c|c|c|c|c|c|c|c|c|}
    \hline
    \multirow{2}{*}{\textbf{Dataset}} &
    \multirow{2}{*}{\textbf{ID}} &
    \multicolumn{4}{c|}{\textbf{Cells and Steps}} &
    \multicolumn{3}{c|}{\textbf{Basic}} &
    \multicolumn{2}{c|}{\textbf{Arch Optim}} &
    \multicolumn{3}{c|}{\textbf{Network Optim}} &
    \multirow{2}{*}{\begin{tabular}[c]{@{}c@{}}\textbf{Model}\\ \textbf{Size}\end{tabular}} &
    \multirow{2}{*}{\begin{tabular}[c]{@{}c@{}}\textbf{Search}\\ \textbf{Time}\end{tabular}} &
    \multirow{2}{*}{\begin{tabular}[c]{@{}c@{}}\textbf{Search}\\ \textbf{Score}\end{tabular}} &
    \multirow{2}{*}{\begin{tabular}[c]{@{}c@{}}\textbf{Eval}\\ \textbf{Score}\end{tabular}} \\ \cline{3-14}
                                                    &   &\textbf{ C}   & \textbf{L}  & \textbf{N} & \textbf{M} & \textbf{Ep} & \textbf{BS} & \textbf{Drpt} & \textbf{LR}   & \textbf{L2}   & \textbf{MaxLR} & \textbf{MinLR} & \textbf{L2}   &   &   &        &        \\ \hline
    NTU                                               & 1 & 128 & 8  & 2 & 2 & 30 & 96 & 0.2  & 3e-4 & 1e-3 & 1e-3  & 1e-6  & 1e-4 & 0.98 M & 53.68 & \textbf{94.48}  & \textbf{90.48}  \\ \hline
    NTU                                               & 2 & 256 & 8  & 2 & 1 & 30 & 96 & 0.2  & 3e-4 & 1e-3 & 1e-3  & 1e-6  & 1e-4 & 1.71 M & 47.76 & 94.16  & 89.19  \\ \hline
    NTU                                               & 3 & 128 & 8  & 2 & 1 & 30 & 96 & 0.2  & 3e-4 & 1e-3 & 1e-3  & 1e-6  & 1e-4 & 0.98 M & 45.84 & 93.01  & 88.78  \\ \hline
    NTU                                               & 4 & 256 & 8  & 4 & 2 & 30 & 96 & 0.2  & 3e-4 & 1e-3 & 1e-3  & 1e-6  & 1e-4 & 2.57 M & 58.64 & 92.22  & 88.30  \\ \hline \hline
    Ego                                               & 1 & 128 & 8  & 2 & 3 & 7  & 72 & 0.0  & 3e-4 & 1e-3 & 3e-3  & 1e-6  & 1e-4 & 0.61 M & 20.67 & \textbf{98.87}  & \textbf{94.96}  \\ \hline
    Ego                                               & 2 & 128 & 8  & 1 & 2 & 7  & 72 & 0.2  & 3e-4 & 1e-3 & 1e-2  & 1e-6  & 1e-4 & 0.45 M & 27.60 & 98.58  & 94.45  \\ \hline
    Ego                                               & 3 & 192 & 8  & 4 & 2 & 7  & 72 & 0.2  & 3e-4 & 1e-3 & 3e-3  & 1e-6  & 1e-4 & 1.17 M & 36.82 & 98.56  & 93.25  \\ \hline
    Ego                                               & 4 & 192 & 12 & 2 & 2 & 7  & 72 & 0.0  & 3e-4 & 1e-3 & 3e-3  & 1e-6  & 1e-4 & 1.59 M & 33.62 & 98.60  & 94.33  \\ \hline \hline
    \begin{tabular}[c]{@{}c@{}}MM\\ IMDB\end{tabular} & 1 & 192 & 16 & 2 & 1 & 12 & 96 & 0.1  & 3e-4 & 1e-3 & 1e-3  & 1e-6  & 1e-4 & 0.65 M & 1.24 & \textbf{53.44} & \textbf{62.92} \\ \hline
    \end{tabular}
    \vspace{-0.2cm}
    \caption{Top-4 Configurations on NTU \cite{shahroudy2016ntu} and EgoGesture \cite{zhang2018egogesture} datasets, and the best configuration on MM-IMDB \cite{arevalo2017gmu} dataset. }
    \vspace{-0.2cm}
    \label{config}
    \end{minipage}
\end{table*}


\section{Search Configurations}

\subsection{Hyper-parameters}

We describe the detailed hyper-parameter configurations on MM-IMDB  \cite{arevalo2017gmu}, NTU RGB-D \cite{shahroudy2016ntu}, and EgoGesture \cite{zhang2018egogesture} datasets in Table \ref{config}, where the notations are discussed in the following.

\noindent\textbf{Cells and steps.}
\textbf{C} is the channels, \textbf{L} is length. In the paper, we refer \textbf{(C, L)} as inner representation size. \textbf{N} is the number of cells, \textbf{M} is the number of steps in each cell. 

\noindent\textbf{Basic training settings.}
\textbf{Ep} is the number of epochs during the searching stage. In the evaluation stage, it could be larger and we roughly set it between $50$ and $70$ in the experiments. \textbf{BS} is the batch size and \textbf{Drpt} is the Dropout rate \cite{srivastava2014dropout}. \textbf{BS} and \textbf{Drpt} is the same for both the searching stage and the evaluation stage.

\noindent\textbf{Architecture optimization.}
For architecture parameter optimization, we use the Adam \cite{kingma2014adam} optimizer. The architecture parameters control the structures of the cells and steps, \ie, $\alpha, \beta, \gamma$ in the paper. \textbf{LR} is the learning rate. \textbf{L2} is the weight decay term.

\noindent\textbf{Network optimization.}
For the network parameters, we use the Adam \cite{kingma2014adam} optimizer with a Cosine Annealing scheduler \cite{loshchilov2016cosineanneal} . The network parameters are trainable parameters from the \emph{fusion network}, including the reshaping layers, cells and the classifier.  \textbf{MaxLR} and \textbf{MinLR} are the learning rate boundaries used by the Cosine Annealing scheduler \cite{loshchilov2016cosineanneal}. \textbf{L2} is the the weight decay term.

\noindent\textbf{Model size and search time.}
\textbf{Model Size} is the total number of parameters of the \emph{fusion network} (in millions), excluding the backbone models. \textbf{Search Time} is the time taken for the searching stage (GPU·hours). We use 8 NVIDIA M40 GPUs in our experiments.

\noindent\textbf{Searching and evaluation scores.}
The \textbf{Search Score} is the performance of the \emph{hypernet} in searching stage on the validation set. The \textbf{Eval Score} is the performance of the \emph{fusion network} in evaluation stage on the test set. For MM-IMDB \cite{arevalo2017gmu} dataset, it includes a multi-label classification task and we use the Weighted F1 score (F1-W) as the metric for performance measurement. For NTU RGB- D\cite{shahroudy2016ntu} dataset and EgoGesture \cite{zhang2018egogesture} dataset, we use the classification accuracy (\%) as the metric.

\newpage

\subsection{Analysis}

To better understand the proposed BM-NAS framework, we empirically study various search configurations of BM-NAS on NTU RGB-D \cite{shahroudy2016ntu} and EgoGesture \cite{zhang2018egogesture}. We list the best configurations we found on NTU RGB-D and EgoGesture in Table \ref{config}. in conjunction with the val/test accuracies. The task on MM-IMDB dataset \cite{arevalo2017gmu} is easier than these two datasets, doesn't require much effort on hyper-parameter tuning, so we only list the best configuration.

Table \ref{config} suggests that $N=2$ might be a good choice. We find that when setting $N=1$, BM-NAS would lean to selecting the late fusion strategy (\ie, selecting the last features of backbones). But as shown in Table 6 in the paper, late fusion may not be the best choice. Regarding the number of steps $M$, $M=2$ already includes many existing fusion strategies (as denoted by Fig. 3 in the paper), while $M=3$ makes a slightly larger search space. We observe that larger $N$ and $M$ may easily lead to overfitting, as there is a total of $N\times M$ inner steps in a Cell. 

\begin{figure}[h]
    \begin{minipage}[b]{1\linewidth}
        \begin{center}
            \includegraphics[width=1\linewidth]{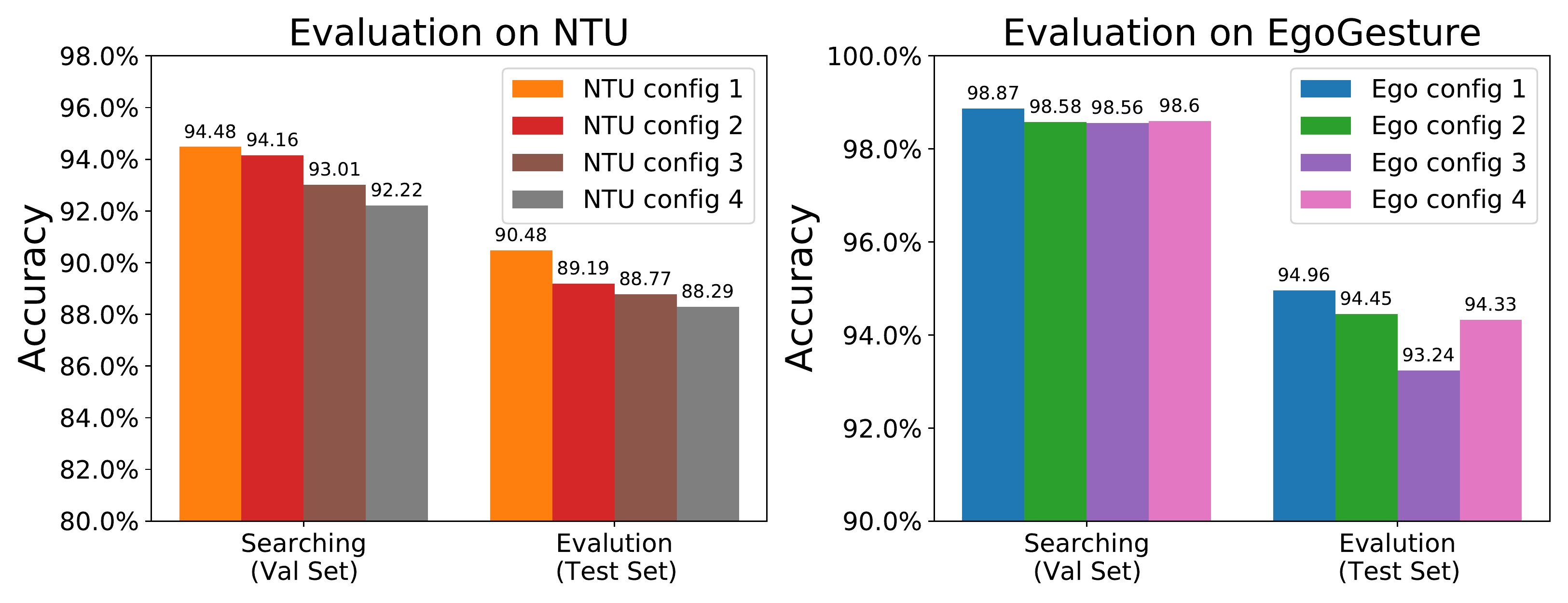}
        \end{center}
        \caption{Performance of hypernets (searching stage) and sampled fusion networks structures (evaluation stage). }
        \label{fig-search-eval-bar}
    \end{minipage}
    \begin{minipage}[b]{1\linewidth}
        \begin{center}
            \includegraphics[width=1\linewidth]{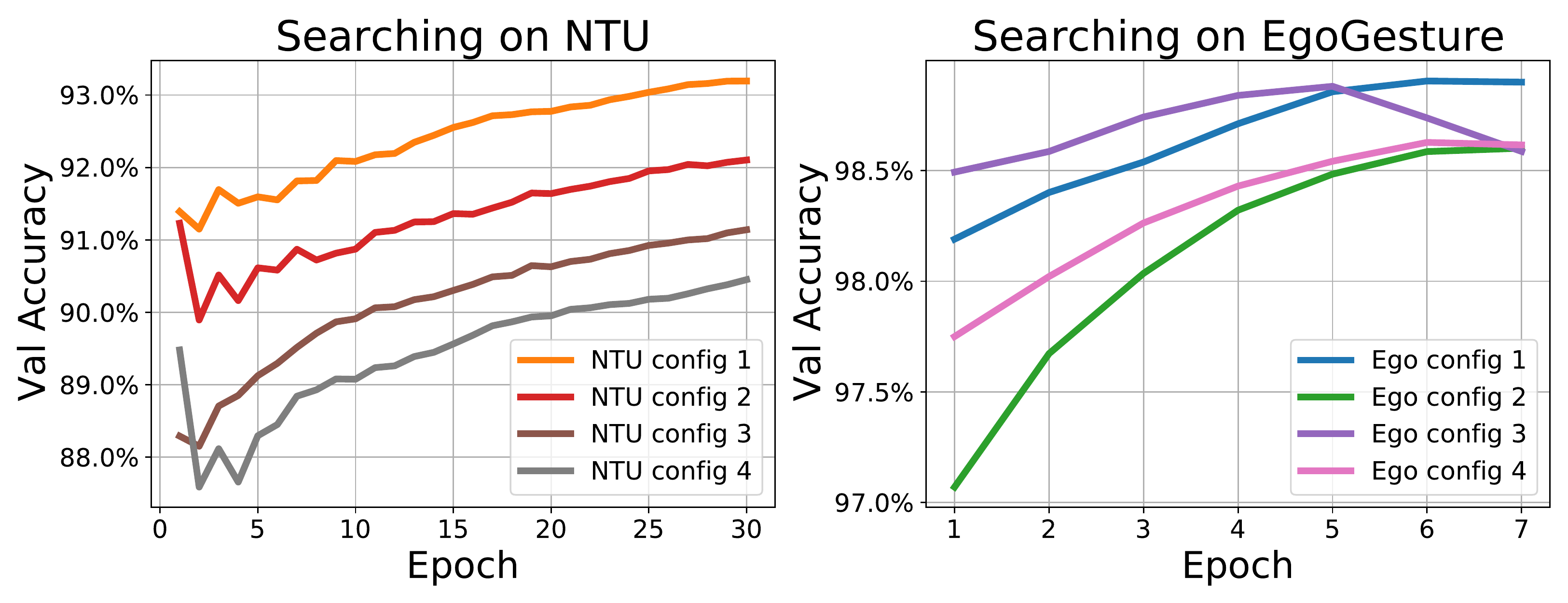}
        \end{center}
        \caption{The validation accuracy of hypernets in searching stage. Results on NTU RGB-D \cite{shahroudy2016ntu} and EgoGesture \cite{zhang2018egogesture} are reported.}
        \label{fig-search-dev-acc-plot}
    \end{minipage}
\end{figure}

With the search configurations in Table \ref{config}, Fig. \ref{fig-search-dev-acc-plot} shows the validation accuracies of the hypernets during search. Fig. \ref{fig-search-eval-bar} further compares the performances of the hypernets, and, compares the performances of the sampled architectures. Figs. \ref{fig-search-dev-acc-plot} and \ref{fig-search-eval-bar} show that the performances of different search configurations of BM-NAS are consistent over searching and evaluation. It suggests that we can select good search configurations according to the validation performance of hypernets instead of performing additional evaluation on the test set with the sampled architecture. 

\newpage

\clearpage

\section{Found Architectures}

\subsection{NTU RGB-D Dataset}

\begin{figure*}[b]
    \centering
    \begin{minipage}{0.9\textwidth}
        \centering
        \subfigure[NTU Config 1]{
            \includegraphics[width=0.45\linewidth]{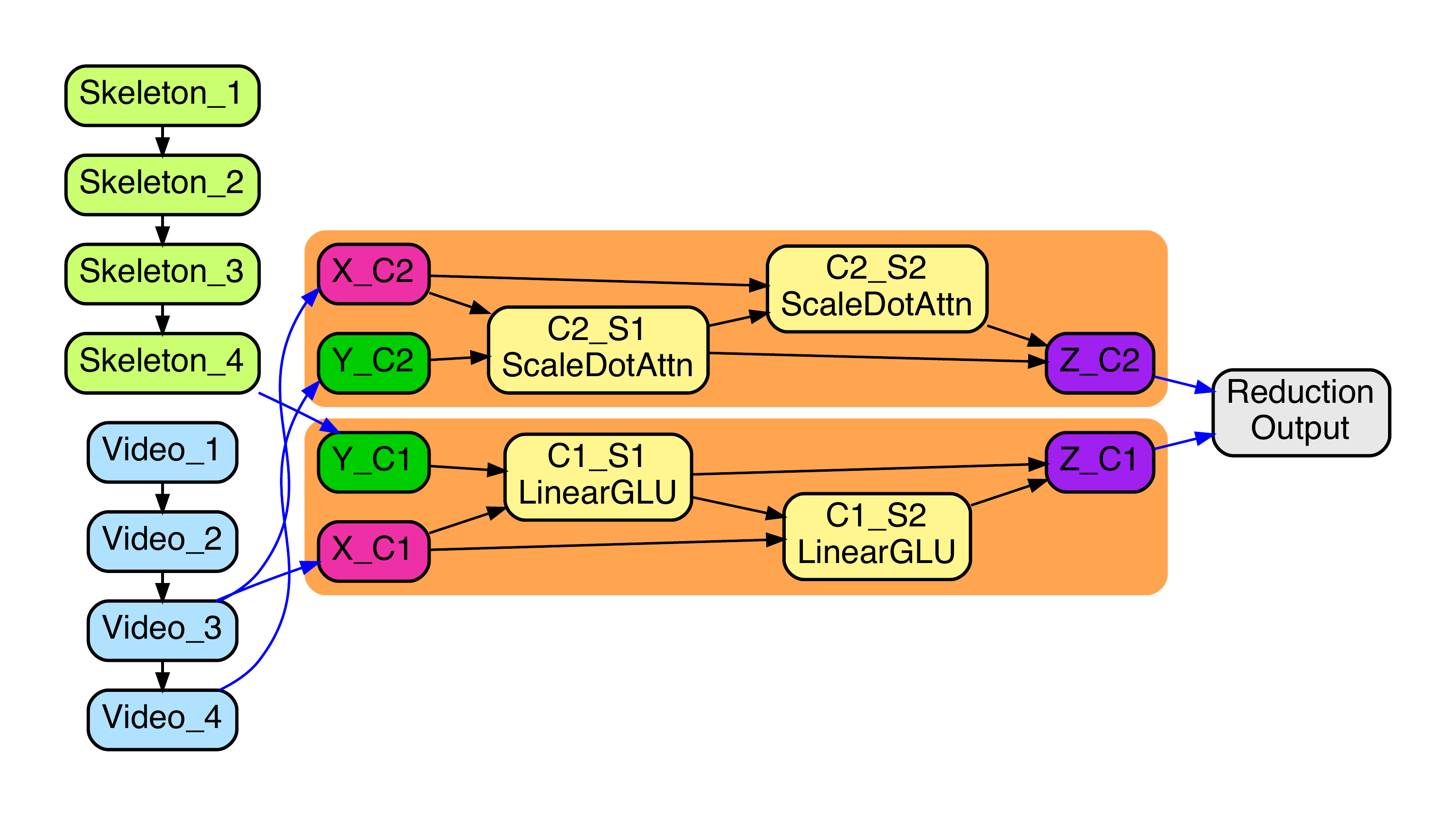}
        }
        \subfigure[NTU Config 2]{
            \includegraphics[width=0.45\linewidth]{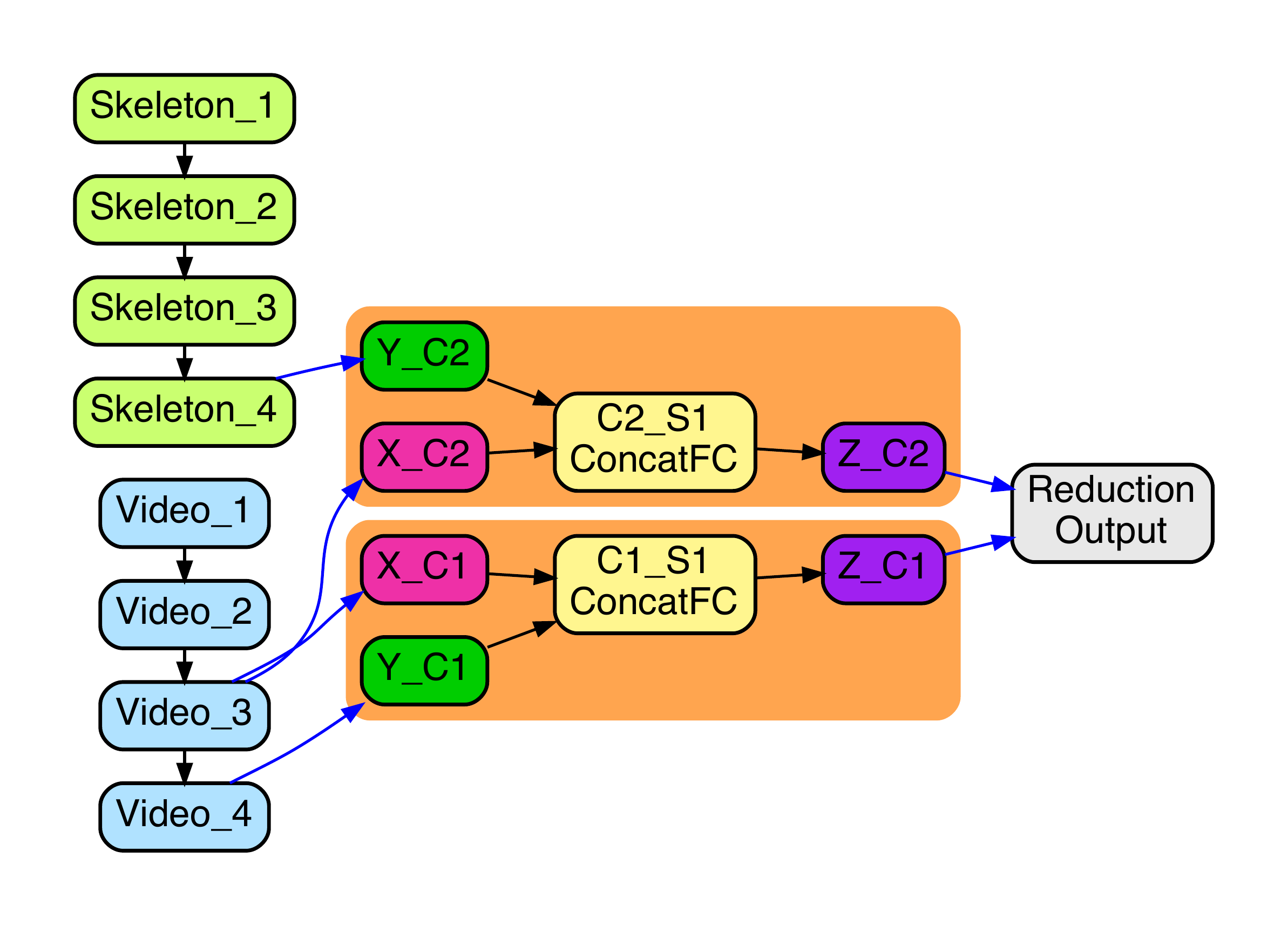}
        }
    \vspace{-0.5cm}
    \end{minipage}
    \begin{minipage}{0.9\textwidth}
        \centering
        \subfigure[NTU Config 3]{
          \includegraphics[width=0.45\linewidth]{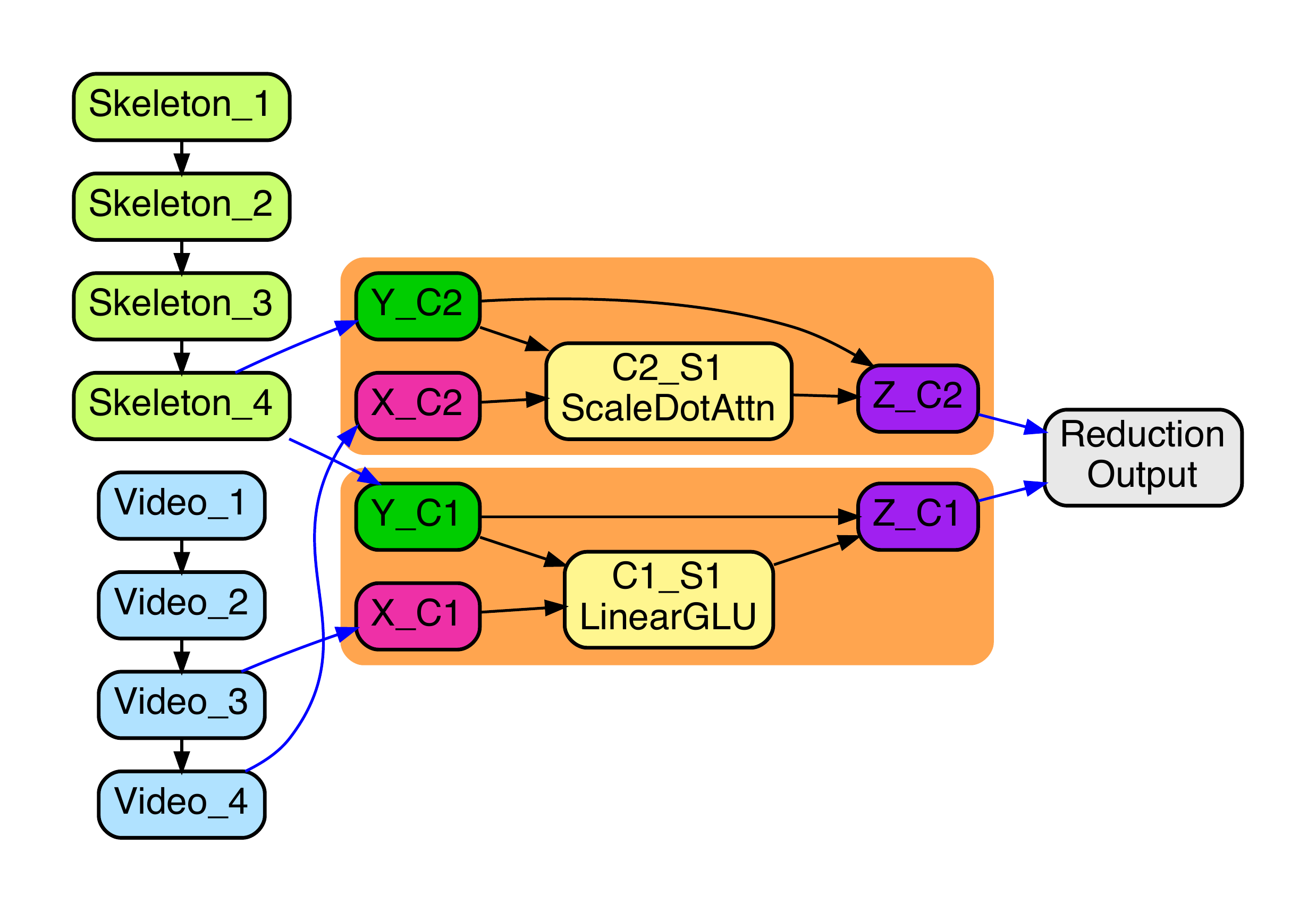}
        }
        \subfigure[NTU Config 4]{ 
          \includegraphics[width=0.45\linewidth]{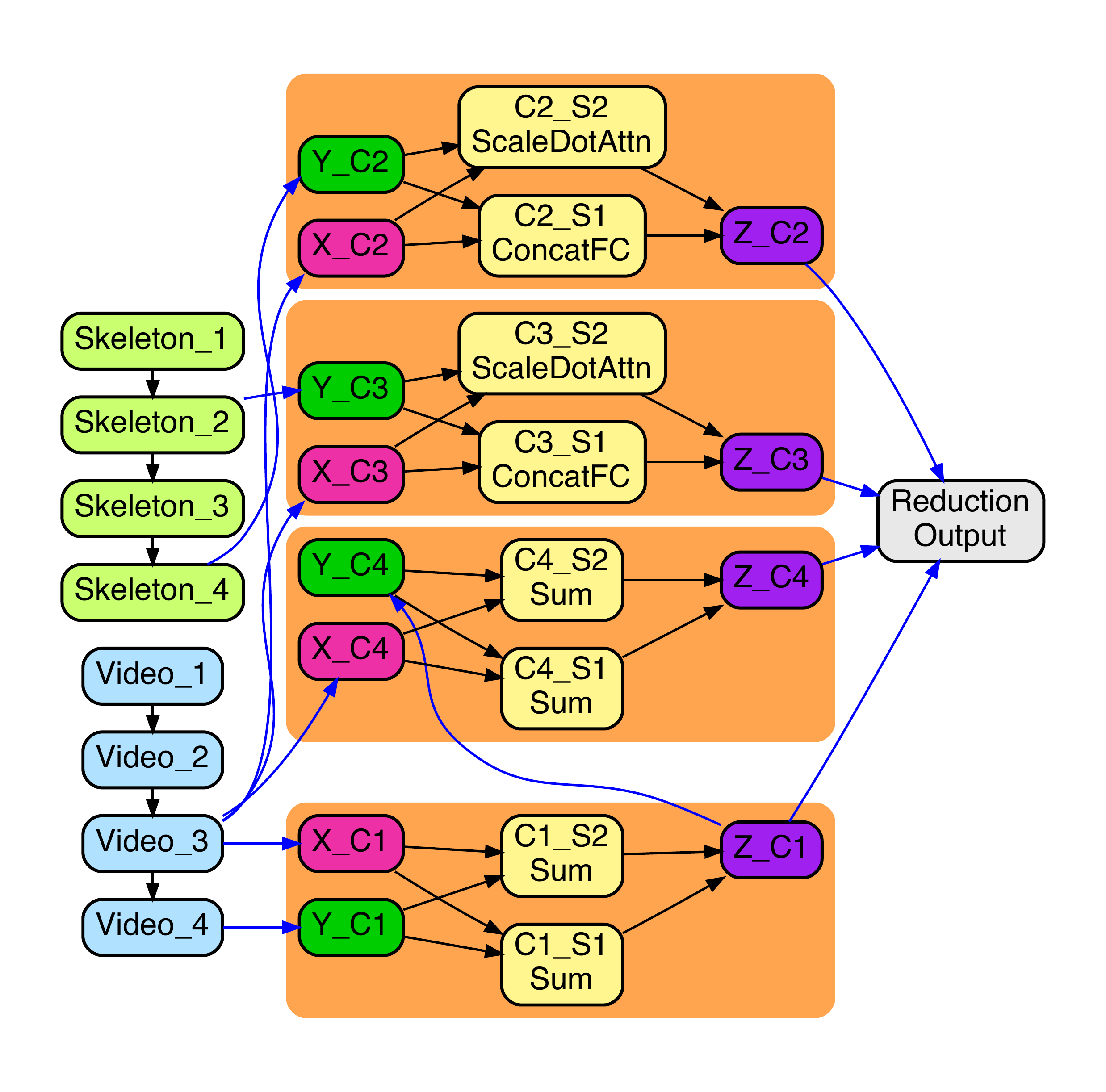}
        }
    \caption{The top-4 architectures found by BM-NAS on NTU \cite{shahroudy2016ntu} dataset. \textbf{`NTU Config 1' is the best architecture found on NTU dataset.} `C1\_S1' denotes $\text{Step}^{(1)}$ of $\text{Cell}^{(1)}$, and so on. The blue edges are the connections at the upper level, and the dark edges are the connections at the lower level.}
    \label{ntu-archs}
    \end{minipage}
\end{figure*}

We tune the hyper parameters extensively on NTU RGB-D \cite{shahroudy2016ntu} dataset. The top-4 configurations are shown in Table \ref{config}, and the architectures found under these configurations are shown in Fig. \ref{ntu-archs}. The `NTU Config 1' is the best architecture found by our BM-NAS framework.

For feature selection strategy, we find that \emph{Video\_3}, \emph{Video\_4}, and \emph{Skeleton\_4} are always selected by our BM-NAS framework no matter how many Cells and steps used. It indicates these are the most effective modality features. Especially \emph{Video\_3} is strongly favored in all the found architectures. MFAS \cite{perez2019mfas} also selects \emph{Video\_4} and \emph{Skeleton\_4} in every found architectures, but it does not pay much attention to \emph{Video\_3}. 

For fusion strategy, we find that adding more inner steps (increasing $M$) is more effective than adding more cells (increasing $N$). However, since we have $N \times M$ steps in total, setting $N$ or $M$ too large would easily lead to an overfitting. Roughly we find that setting $N=2$, $M=2$ is a good option. $N=2$ means we have two different feature pairs for Cells, which is sufficient to cover the three most important features \emph{Video\_3}, \emph{Video\_4} and \emph{Skeleton\_4}. And $M=2$ is sufficient for BM-NAS to form all the fusion strategy like concatenation, attention on attention (AoA) \cite{huang2019aoa}, \etc, as shown in the paper. The best fusion strategy found by BM-NAS on NTU is very similar to an AoA \cite{huang2019aoa} module, see `NTU Config 1' in Fig. \ref{ntu-archs}.

\clearpage

\subsection{EgoGesture Dataset}

\begin{figure*}[b]
    \centering
    \begin{minipage}{1\textwidth}
        \centering
        \subfigure[Ego Config 1]{
            \includegraphics[width=0.48\linewidth]{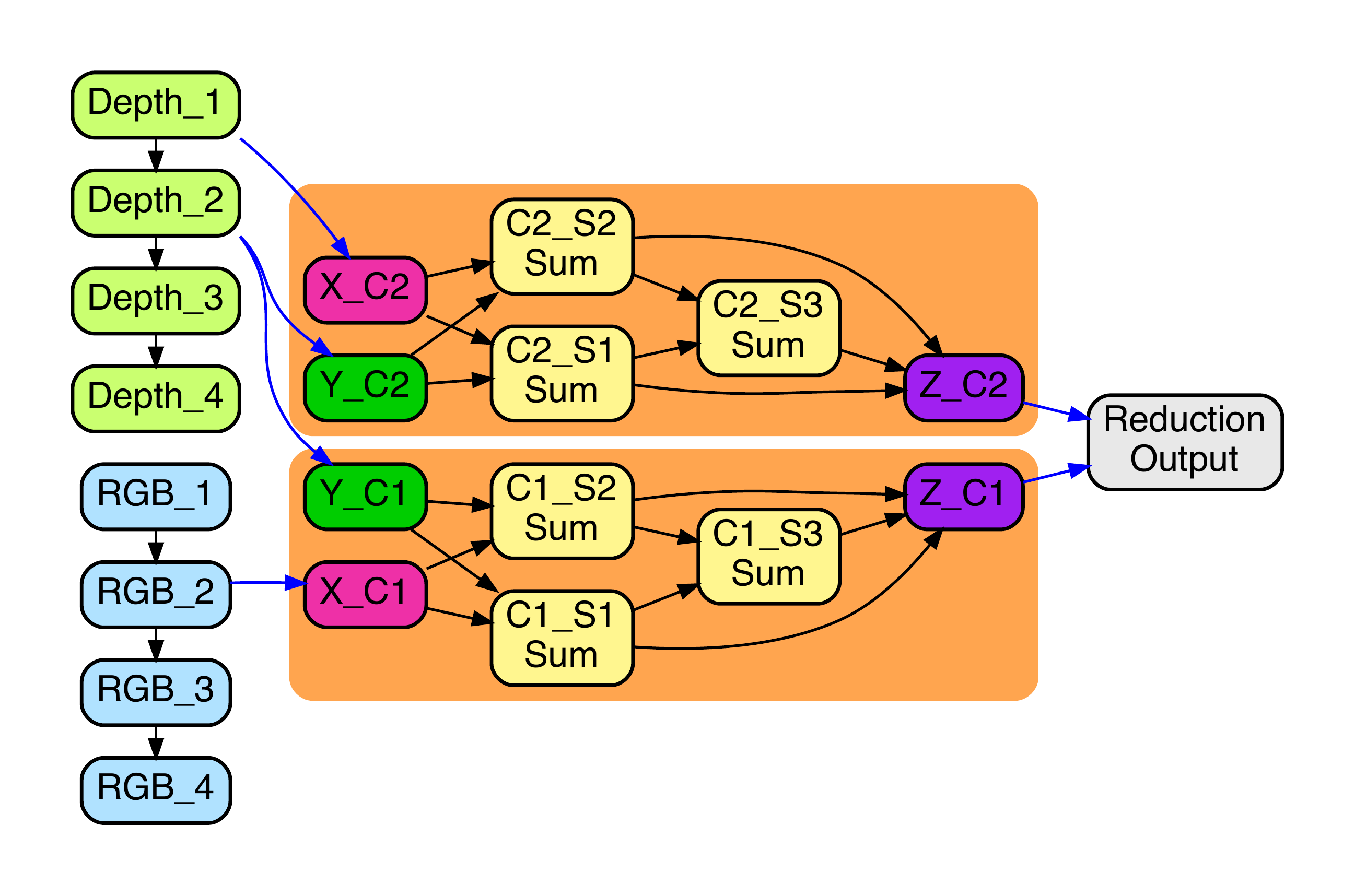}
        }
        \subfigure[Ego Config 2]{
            \includegraphics[width=0.48\linewidth]{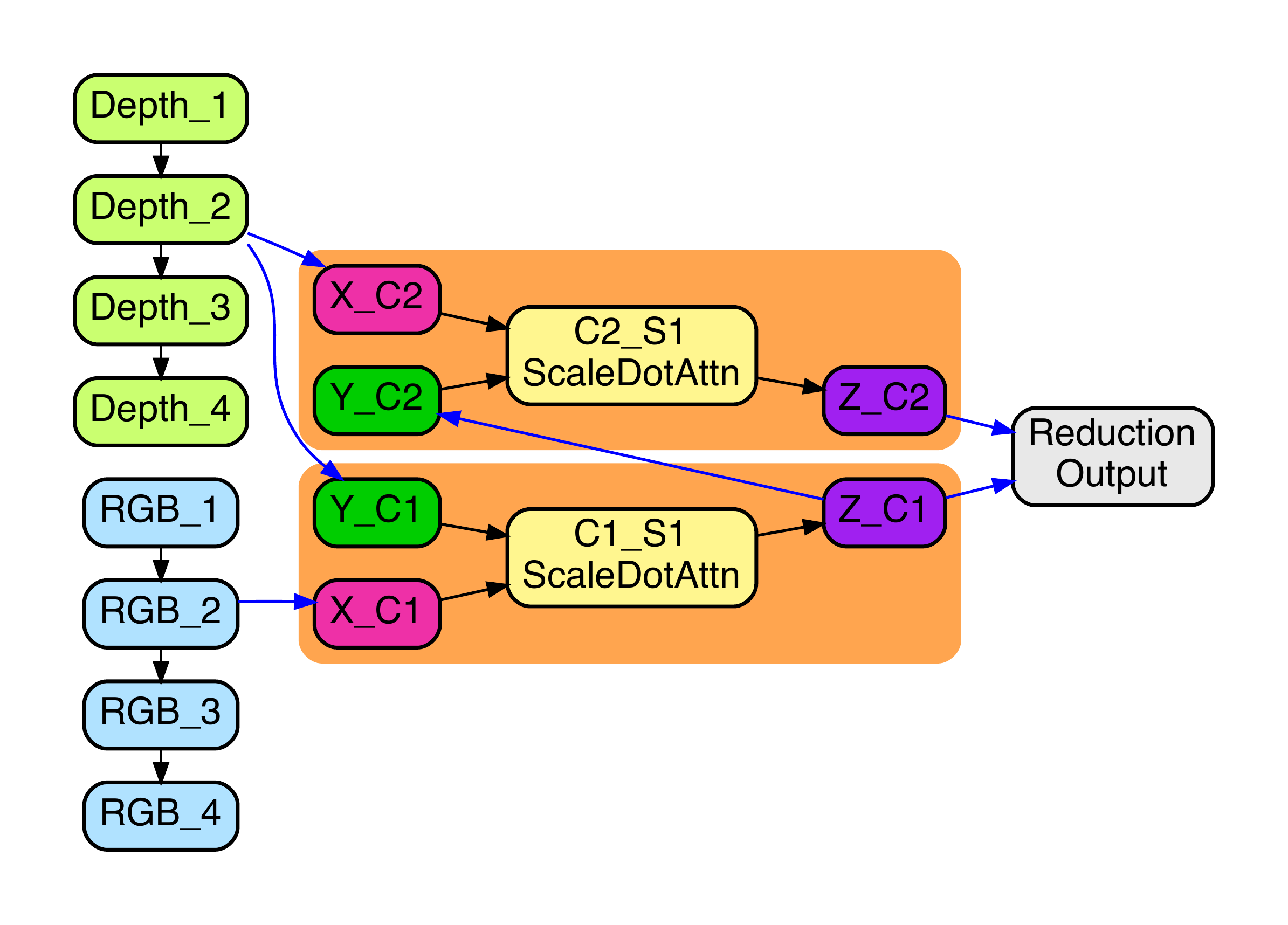}
        }
    \vspace{-0.5cm}
    \end{minipage}
    \begin{minipage}{1\textwidth}
        \centering
        \subfigure[Ego Config 3]{
            \includegraphics[width=0.48\linewidth]{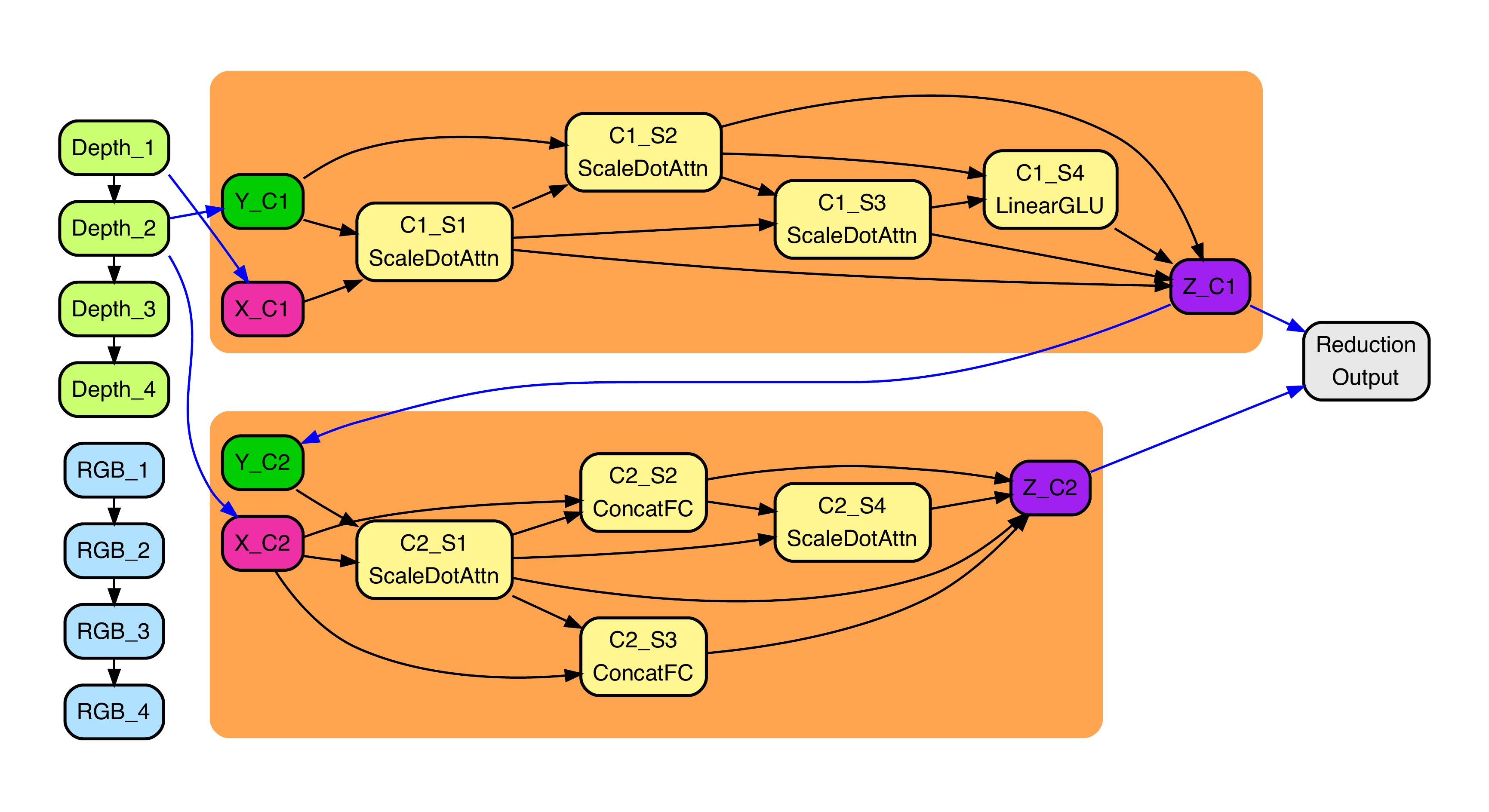}
        }
        \subfigure[Ego Config 4]{
            \includegraphics[width=0.48\linewidth]{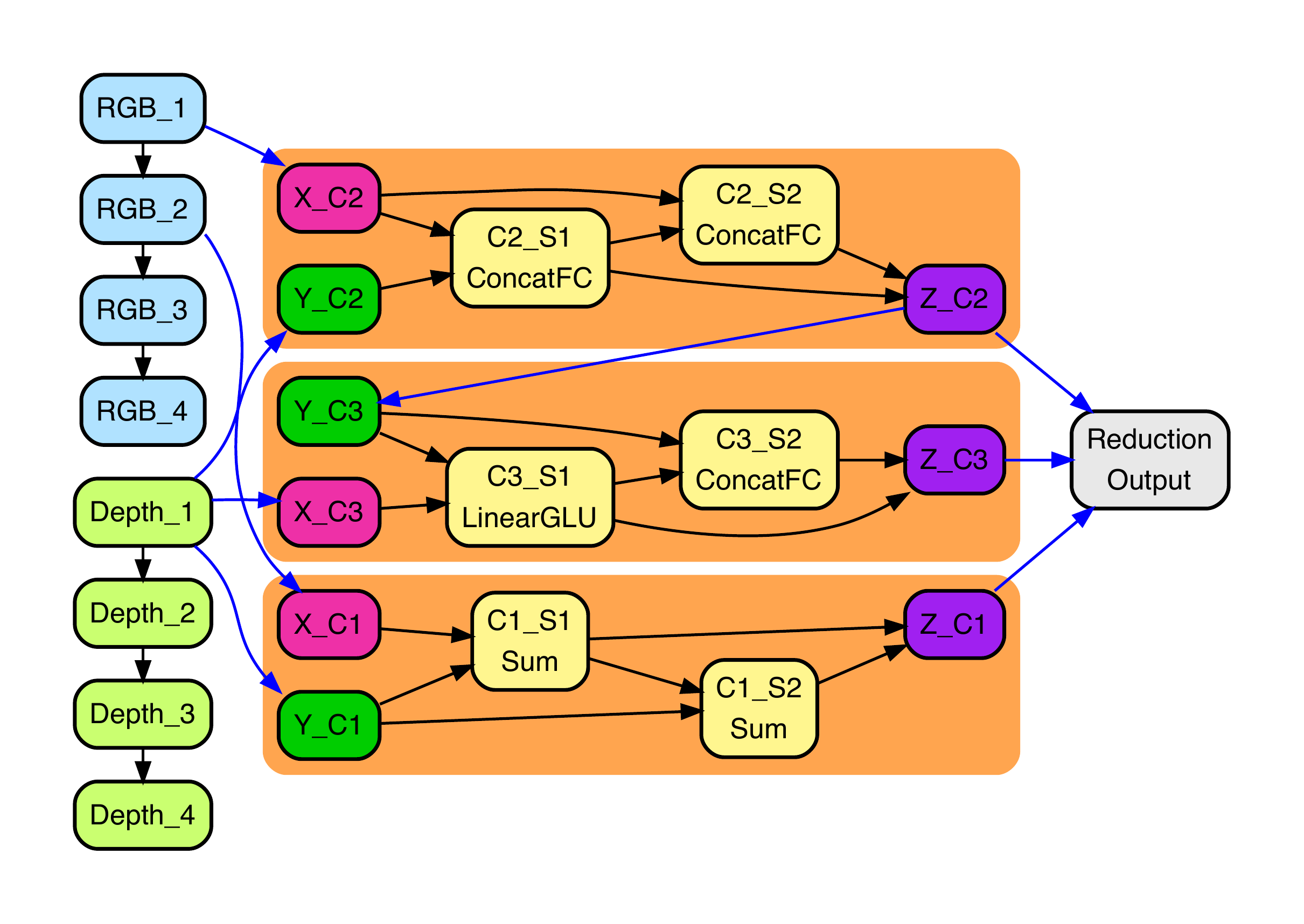}
        }
    \caption{The top-4 architectures found by BM-NAS on EgoGesture\cite{zhang2018egogesture} dataset. \textbf{`Ego Config 1' is the best architecture found on EgoGesture dataset.} }
    \label{ego-archs}
    \end{minipage}
\end{figure*}

For the experiments on EgoGesture \cite{zhang2018egogesture} dataset,
we basically follow the settings as those in NTU RGB-D dataset. The top-4 configurations are shown in Table \ref{config}, and the architectures found under these configurations are shown in Fig. \ref{ego-archs}. The `Ego Config 1' is the best architecture found by our BM-NAS framework.

For feature selection strategy, we find \emph{Depth\_1}, \emph{Depth\_2}, and \emph{RGB\_2} are the most important features for EgoGesture \cite{zhang2018egogesture}. 

For fusion strategy, we find that a combination of $\mathrm{Sums}$ is the most effective, shown as `Ego Config 1' in Fig. \ref{ego-archs}, probably because the backbone models share the same architecture. Unlike the experiments on NTU RGB-D \cite{shahroudy2016ntu} which use Inflated ResNet-50 \cite{baradel2018inflated} for RGB videos and Co-occurrence \cite{li2018co} for skeletons modality, EgoGesture \cite{zhang2018egogesture} uses ResNeXt-101 \cite{kopuklu2019real} backbone for both the depth videos and the RGB videos. These two backbone models have exactly the same architecture, except for the input channels of the first convolutional layer. Therefore, the depth features and the RGB features probably share the semantic levels for features of the same depths, such as \emph{Depth\_2} and \emph{RGB\_2} in `Ego Config 1'. 

\newpage

\subsection{MM-IMDB Dataset}

\begin{figure}[h]
    \begin{center}
    \includegraphics[width=1\linewidth]{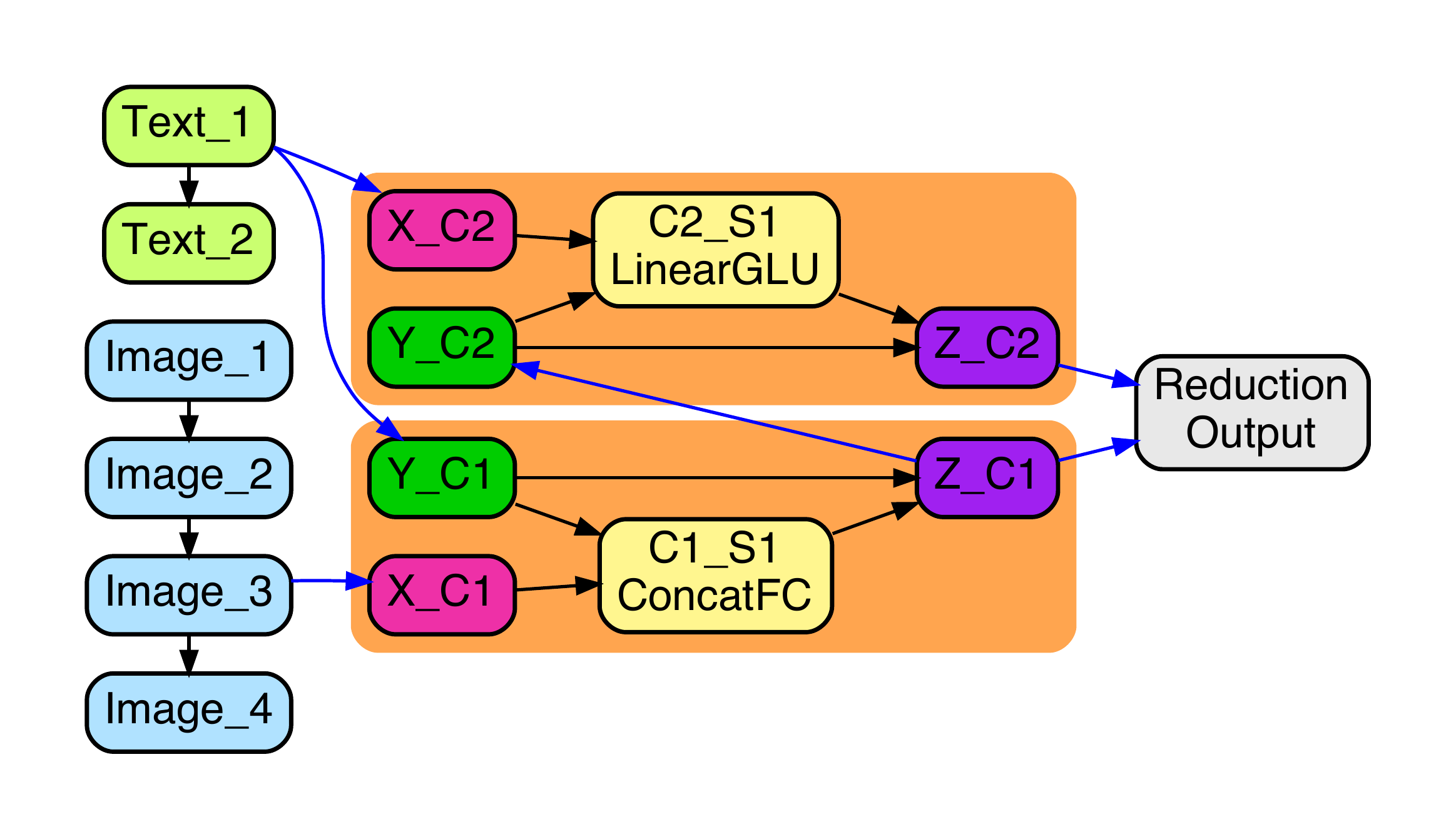}
    \end{center}
    \caption{MM-IMDB Config 1, which is the best architecture found by BM-NAS on MM-IMDB \cite{arevalo2017gmu} dataset. }
    \label{fig-best-mmimdb-model}
\end{figure}

We do not tune the hyper-parameters extensively on MM-IMDB \cite{arevalo2017gmu} since it is a relatively simple task compared with NTU RGB-D \cite{shahroudy2016ntu} and EgoGesture \cite{zhang2018egogesture}. The configuration can be found in Table \ref{config}. As shown in Fig. \ref{fig-best-mmimdb-model}, we find \emph{Image\_2} and \emph{Text\_0} are the most important modality features. The best fusion operation is \emph{ConcatFC} for \emph{Image\_2} and \emph{Text\_0}, and \emph{LinearGLU} for \emph{Cell\_0} and \emph{Text\_0}.

It is worth noting that we use the Weighted F1 score (F1-W) as the metric for performance, since we perform a multi-label classification task on MM-IMDB\cite{arevalo2017gmu} dataset. Although the Macro F1 score (F1-M) is also reported in the paper, we only use F1-W for model selection, because the distribution of labels in MM-IMDB \cite{arevalo2017gmu} is highly imbanlanced as illustrated in Table \ref{mmimdb-labels}. Thus, F1-W would be a better metric as F1-M does not take label imbalance into account. 

\begin{table}[h]
\begin{minipage}{1\linewidth}
\centering
    \begin{tabular}{|c|c|c|c|}
        \hline
        \textbf{Label}       & \textbf{\#Samples} & \textbf{Label}      & \textbf{\#Samples} \\ \hline
        Drama       & 13967  & War        & 1335   \\ \hline
        Comedy      & 8592   & History    & 1143   \\ \hline
        Romance     & 5364   & Music      & 1045   \\ \hline
        Thriller    & 5192   & Animation  & 997    \\ \hline
        Crime       & 3838   & Musical    & 841    \\ \hline
        Action      & 3550   & Western    & 705    \\ \hline
        Adventure   & 2710   & Sport      & 634    \\ \hline
        Horror      & 2703   & Short      & 471    \\ \hline
        Documentary & 2082   & Film-Noir  & 338    \\ \hline
        Mystery     & 2057   & News       & 64     \\ \hline
        Sci-Fi      & 1991   & Adult      & 4      \\ \hline
        Fantasy     & 1933   & Talk-Show  & 2      \\ \hline
        Family      & 1668   & Reality-TV & 1      \\ \hline
        Biography   & 1343   &            &        \\ \hline
        \end{tabular}
        \vspace{0.5cm}
        \caption{Label distribution of MM-IMDB\cite{arevalo2017gmu} dataset.}
        \label{mmimdb-labels}

\end{minipage}

\end{table}

\clearpage
\bibliography{aaai22}
\end{document}